\documentclass[sensors,article,moreauthors,pdftex,10pt,a4paper]{mdpi} 

\firstpage{1} 
\pubyear{2018}
\copyrightyear{2018}

\usepackage{amssymb,amsfonts}
\usepackage[all]{myglossary}

 \theoremstyle{mdpi}
 \newcounter{thm}
 \newcounter{ex}
 \newcounter{re}
 \theoremstyle{mdpidefinition}

\Title{Extending QGroundControl for automated mission planning of UAVs}

\Author{Cristian Ramirez-Atencia and David Camacho}
\AuthorNames{Cristian Ramirez-Atencia and David Camacho}

\address{%
Universidad Aut\'onoma de Madrid, 28049 Madrid, Spain; cristian.ramirez@inv.uam.es (C.R.A.); david.camacho@uam.es (D.C.)}

\corres{Correspondence: cristian.ramirez@inv.uam.es}

\preto{\abstractkeywords}{\nolinenumbers}

\abstract{\glspl*{uav2} have become very popular in the last decade due to some advantages such as strong terrain adaptation, low cost, zero casualty and so on. One of the most interesting advances in this field is the automation of mission planning (task allocation) and real-time replanning, which are highly useful to increase the autonomy of the vehicle and reduce the operator workload. These automated mission planning and replanning systems require a \gls*{hci} that facilitates the visualization and selection of plans that will be executed by the vehicles. In addition, most missions should be assessed before their real-life execution. This paper extends QGroundControl, an open-source simulation environment for flight control of multiple vehicles, by adding a mission designer, that permits the operator to build complex missions with tasks and other scenario items; an interface for automated mission planning and replanning, which works as test bed for different algorithms, and a \gls*{dss} that helps the operator in the selection of the plan. In this work, a complete guide of these systems and some practical use cases are provided.}


\begin{document}

\nolinenumbers

\glsresetall

\section{Introduction}
\label{sec:introduction}
The use of \glspl*{uav2}, also referred to as drones, has highly increased in the last decade, becoming very popular in many applications including traffic monitoring~\cite{Kanistras2015Survey}, agriculture~\cite{Zhang2012Application} or disaster and crisis management~\cite{Erdelj2016UAV}, since they avoid risking human lives while their manageability permits to reach areas of hard access. These vehicles are usually controlled by a number of operators inside one or more \glspl*{gcs}, depending on the size of the mission.

Automated mission planning over a swarm of \glspl*{uav2} remains to date as a challenging research trend in what regards to this particular type of aircrafts. This problem involves generating tactical goals, commanding vehicles, risk avoidance, coordination and timing. Currently, \glspl*{uav2} are controlled remotely by human operators using rudimentary planning systems, such as pre-configured plans, classical planners that are not able to cope with the entire complexity of the problem, or manually provided schedules. Some recent works~\cite{Evers2014Robust,Ramirez2018Weighted} have provided more efficient approaches to solve the \gls*{mcmpp} considering several features of the problem such as time constraints, fuel constraints, sensor constraints, etc. Due to its complexity and multiple conflicting criteria (e.g. makespan, cost or risk of the mission), multi-objective solvers such as \glspl*{moea} have been used in these works.

One of the most challenging problems in this field is mission replanning, which implies a new planning for the previous mission plan due to certain incidences, such as a vehicle or sensor failure or a new task arrival, during the real-time execution of the mission. A few recent works have developed systems that deal with automated mission replanning, based on a repair of the previous plan~\cite{Fukushima2012Onboard}, or performing a full replanning of the mission in a limited runtime~\cite{Ramirez-Atencia2016MOGAMR}.

Due to the complexity and multiple conflicting objectives of this problem, several non-dominated solutions (i.e. the \gls*{pof}) are obtained. This situation hinders the process of decision making for the operator when selecting the final plan. In order to reduce his/her workload, a \gls*{dss} can be provided to help the operator in the plan selection. This \gls*{dss} can work in two steps: first, inside the mission planning algorithm, focusing the search of solutions on the most relevant ones, which can be made using a knee point based \gls*{moea}~\cite{Ramirez-Atencia2017Knee}. Secondly, once the most relevant solutions are returned, the solutions are ranked using some \gls*{mcdm} technique based on the operator preferences, and filtered based on the similarity of the obtained solutions.

All of these techniques developed to solve the mission planning and replanning problems, must be properly tested by expert operators in a simulated environment before they are considered apt for real \gls*{uav2} missions. In this work, QGroundControl~\cite{qgroundcontrol}, an open-source ground control station simulator, has been extended by adding an automated planning interface, so this framework can work as a test bed for mission planning and replanning algorithms, and also for decision making methods. This extension allows operators to automatically plan a mission, simulate this plan and then perform a replanning during the execution.

\begin{figure}[H]
\centering
\includegraphics[width=0.9\textwidth]{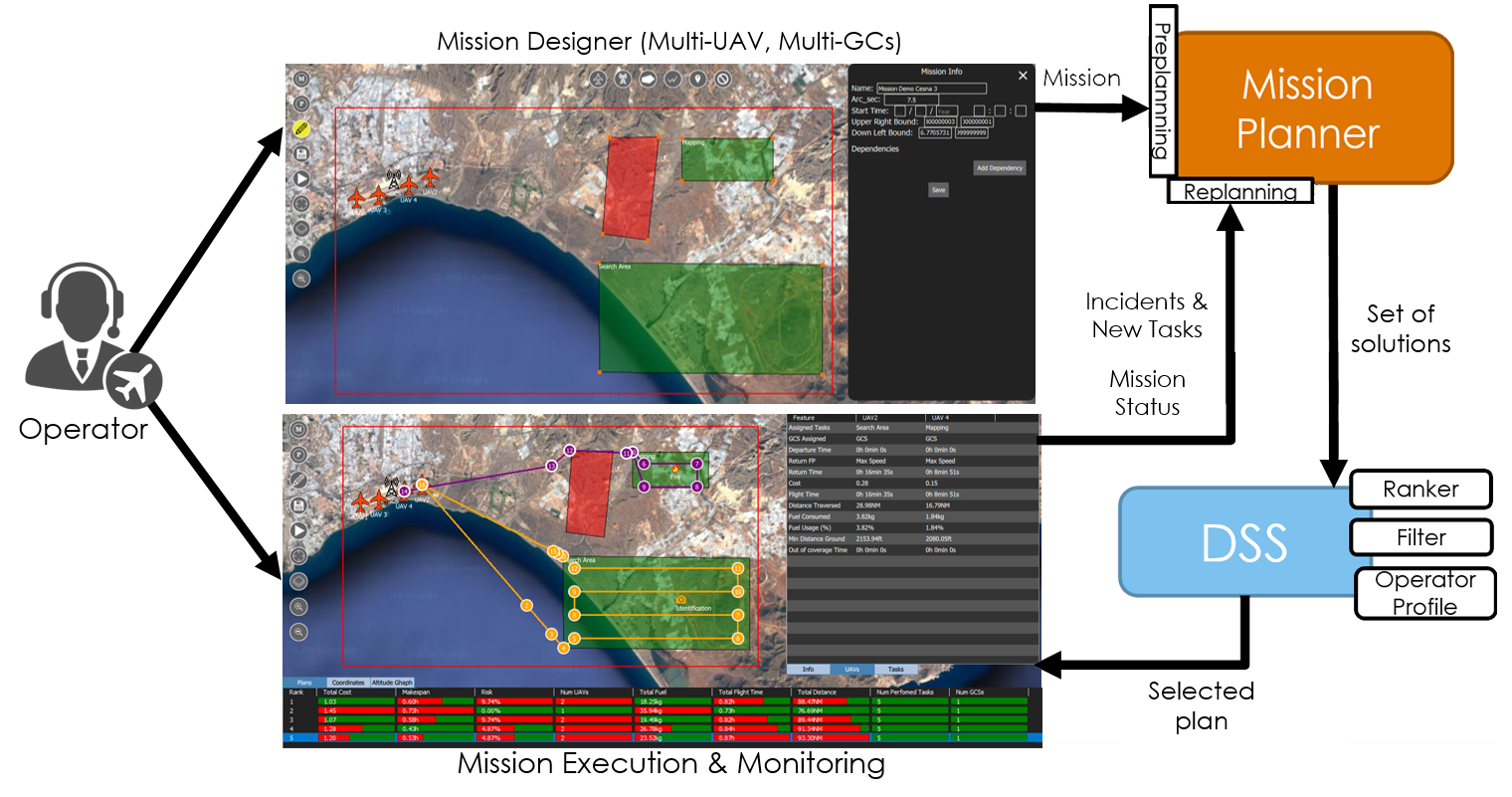}
\caption{Architecture of the framework extended from QGroundControl, including mission (re)planning and decision support.}\label{fig:architecture}
\end{figure}

Additionally, in order to ease the entry of the definition of the mission and the scenario, a graphical mission designer has been built inside \textit{QGroundControl}. This designer permits to create a mission with all its elements (\glspl*{uav2}, tasks, \glspl*{gcs}, \glspl*{nfz}, ...). After that, the mission can be automatically planned and the generated plans can be visualized. Finally, one of these plans can be executed and the \glspl*{uav2} are monitored all together. The architecture of the proposed framework is presented in Figure \ref{fig:architecture}. This extended tool is not publicly accessible by the moment due to confidential issues\footnote{If the reader is interested in testing the current version of the framework, please contact with the corresponding author.}.

The following sections are as follows: Section \ref{relatedwork} provides a background and related works on ground control station frameworks and test bed interfaces. Section \ref{designer} describes the mission designer and its components. Section \ref{planner} explains the mission planning problem, and how the automated mission planner and the \gls*{dss} have been integrated within QGroundControl. Section \ref{replanning} explains the mission replanning problem and how it has been integrated in the simulation environment. Section \ref{casestudy} presents some use cases that has been performed to prove the functionalities developed by creating, planning, simulating and replanning a mission. Finally, Section \ref{conclusions} draws some conclusions and future work.


\section{Related work}\label{relatedwork}
In the last years, some works have proposed \gls*{uav2} simulation environments for supervisory flight control~\cite{Sibley2016Supporting},~coordination \cite{Garcia2010Multi} or training~\cite{Rodriguez-Fernandez2015Design}.

When working with simulation and control of \glspl*{uav2}, there are two type of software that must be differentiated: the autopilot software and the \gls*{gcs} software. The autopilot controls automatically the trajectory of the \gls*{uav2}, and can provide the telemetry of the vehicle. The most known autopilot simulators are ArduPilot~\cite{ardupilot} and PX4~\cite{px4}.

On the other hand, \gls*{gcs} software focuses on the operator side, providing flight control and manual path planning of one or multiple vehicles. In order to communicate these \glspl*{gcs} with the autopilots, a communication protocol is required. The most used protocol, able to provide communication with both ArduPilot and PX4 is MAVLink~\cite{mavlink}. This protocol is used in the most known \gls*{gcs} tools, such as \textit{MAVProxy}~\cite{mavproxy}, \textit{Mission Planner}~\cite{missionplanner}, \textit{APM Planner 2}~\cite{apmplanner2}, \textit{UgCS}~\cite{ugcs} and QGroundControl~\cite{qgroundcontrol}. \textit{QGroundControl} is the only one of them that permits the control of multiple \glspl*{uav2} simultaneously, although \textit{UgCS} provides a much more proficient interface with many features such as \glspl*{nfz} and immersive 3D simulation.

One of the most popular open-source software for small drones is \textit{Paparazzi UAV}~\cite{paparazzi}, which provides both the autopilot and the \gls*{gcs} tools. It is a very complete framework and also allows control of multiple vehicles simultaneously.

\textit{QGroundControl} is open source and provides full ground station support and flight control and configuration for multiple \glspl*{uav2} through MAVLink communications, allowing to control both ArduPilot and PX4 vehicles. The main power of \textit{QGroundControl} is that it provides easy and straightforward usage for beginners, while still delivering high end feature support for experienced users. It has a easy-use path planning interface (through waypoint insertion) for autonomous flight. It also allows flight map display showing vehicle position, flight track, waypoints and vehicle instruments, and video streaming.

Nevertheless, QGroundControl, as well as the rest of \gls*{gcs} software, only permits to create manual plans by waypoint insertion for each \gls*{uav2}, i.e. no automated planning algorithm is provided within the framework. It does not allow to create tasks nor \glspl*{nfz} in order to create a mission which could be automatically planned. On the other hand, QGroundControl only permits to see the waypoint plan of one \gls*{uav2} at a time, so for multi-UAV missions it is complicated to monitor all vehicles at once.

A test bed interface (which has been used in many works specially for providing different artificial intelligence algorithms for games~\cite{Buro2003RTS}, and also for flight control~\cite{Oliveira2011Test}) for automated mission planning and replanning is a novel requirement that has not been so far implemented inside \glspl*{gcs}. This interface must allow through a communications protocol, the use of different automated mission planning and replanning algorithms. This paper provides an extension of \textit{QGroundControl} providing this test bed and all the lacked capabilities mentioned.


\section{Mission Designer}\label{designer}
A mission is composed of a set of \textit{objectives} to be performed by a swarm of \glspl*{uav2}. These objectives are composed by one or more \textit{tasks}, where a task is indivisible and must be performed within a specific time interval using some sensors carried by the vehicle. These tasks may have some dependencies between them, restricting the order of the tasks. Each mission should be performed in a specific geographic zone or scenario, where there could be some \glspl*{nfz} that must be avoided by the vehicles. In addition, one or more \glspl*{gcs} control the swarm of \glspl*{uav2}.

\begin{figure}[H]
\centering
\includegraphics[width=0.9\textwidth]{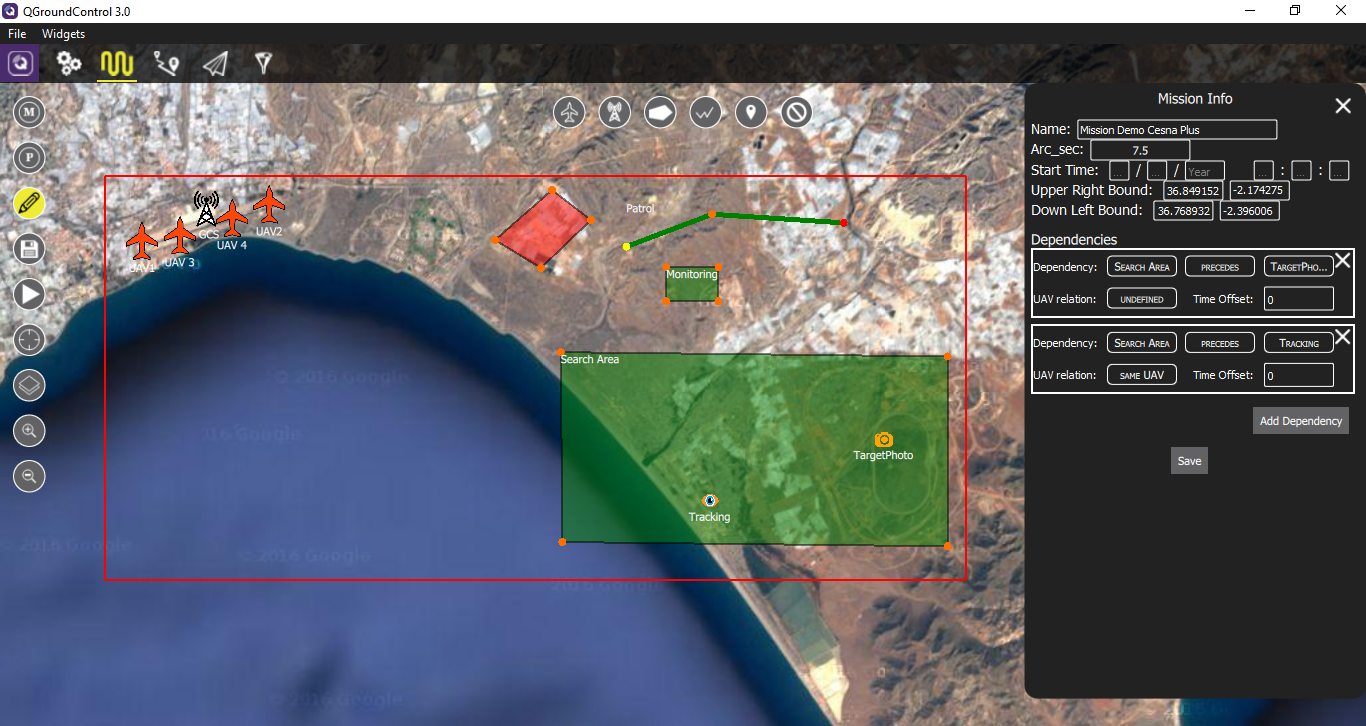}
\caption{Mission Designer in the QGroundControl for adding new elements.}\label{fig:mission}
\end{figure}

In order to permit the operators to create new missions and fulfill them with all the elements involved in the mission (tasks, vehicles, \glspl*{gcs}, ...), a mission designer has been developed (see Figure \ref{fig:mission}). The integration of these new functionalities into QGroundControl requires the modification of this software in order to extend its functionality. For this, a new tab has been added in the main menu of the QGroundControl (represented in Figure \ref{fig:mission} with a looped path in the top bar). This tab is used both for the Mission Designer and the rest of functionalities added in the following sections.

The mission designer permits to create new missions or read already created ones, and provides a set of tools for adding different elements to the mission scenario. In the following subsections, we explain how to create a new mission and how to add the different elements to it using the mission designer (represented with a pencil in Figure \ref{fig:mission}).

\subsection{Creating new missions}\label{create}
To create a new mission, there are several parameters of the mission that must be provided by the operator:

\begin{itemize}[leftmargin=*,labelsep=4mm]
\item The name of the mission: This name will be used later to identify this mission when reading it.
\item The bounds of the mission scenario: The upper-right and down-left bounds that identify the limits of the mission, given in latitude and longitude degrees.
\item The arc-seconds of the elevation map to be used by the mission planner. The possible values are $30$, $15$ and $7.5$ arc-seconds.
\item The start time of the mission (optional), expressed as date and time. If not provided, the time will be taken from the CPU clock when the simulation starts.
\end{itemize}

After fulfilling these values, the mission will be created and an empty scenario inside the defined bounds will be presented.

\subsection{Adding a new UAV}\label{adduav}
To add a \gls*{uav2}, the first button in the top of the scenario, represented as a plane in Figure \ref{fig:mission}, is used. When adding a new vehicle, some properties must be considered:

\begin{itemize}[leftmargin=*,labelsep=4mm]
\item The name of the \gls*{uav2}.
\item The initial amount of fuel of the \gls*{uav2}, expressed in $Kg$.
\item The position of the \gls*{uav2} (latitude, longitude and altitude in $ft$).
\item The (optional) end position for the departure runway of the \gls*{uav2}, to where it must go when taking off.
\item The (optional) start position for the landing track of the \gls*{uav2}, where it must first go when landing.
\item The (optional) end position of the \gls*{uav2}, where it must land when the mission ends.
\item The (optional) start and end times of use of the \gls*{uav2}. These are expressed as a date and time, and can only be used when the start time of the mission has been defined.
\item The type of the \gls*{uav2}. This property defines the type of vehicle used (e.g. HALE, MALE, URAV, UCAV, ...). This property comprises a set of characteristics of the \gls*{uav2}:

\begin{itemize}[leftmargin=*,labelsep=4mm]
\item The mass of the vehicle (in Kg).
\item The maximum fuel capacity (in Kg).
\item The cost per hour.
\item The maximum altitude limit (in ft).
\item The maximum speed limit (in knots).
\item The maximum flight time (in hours).
\item The maximum range or distance (in NM).
\item A set of flight profiles, defining the performance of the vehicle in terms of speed (in knots), fuel consumption (in Kg/h) and altitude (in ft) or angle of climb/descent (in degrees). In this work, the flight profiles considered for every vehicle are a minimum consumption profile, a maximum speed profile, a climb profile and a descent profile.
\end{itemize}

\item The configuration of the \gls*{uav2}. This property defines the configuration of the vehicle, specifically, the set of sensors carried by it.
\end{itemize}

\subsection{Adding a new GCS}\label{addgcs}
This function is represented by the second button in the top of the scenario in Figure \ref{fig:mission}. When adding a new \gls*{gcs}, some properties must be fulfilled:

\begin{itemize}[leftmargin=*,labelsep=4mm]
\item The name of the \gls*{gcs}.
\item The position of the \gls*{gcs} (latitude, longitude and altitude in $ft$).
\item The type of \gls*{gcs}. This property defines the type of station used, which comprises a set of characteristics:
\begin{itemize}[leftmargin=*,labelsep=4mm]
\item The within range of communications (in NM).
\item The maximum number of vehicles that the station can control simultaneously.
\item The type of vehicles that the station can control.
\end{itemize}
\end{itemize}

\begin{figure}[t]
\centering
\includegraphics[width=0.9\textwidth]{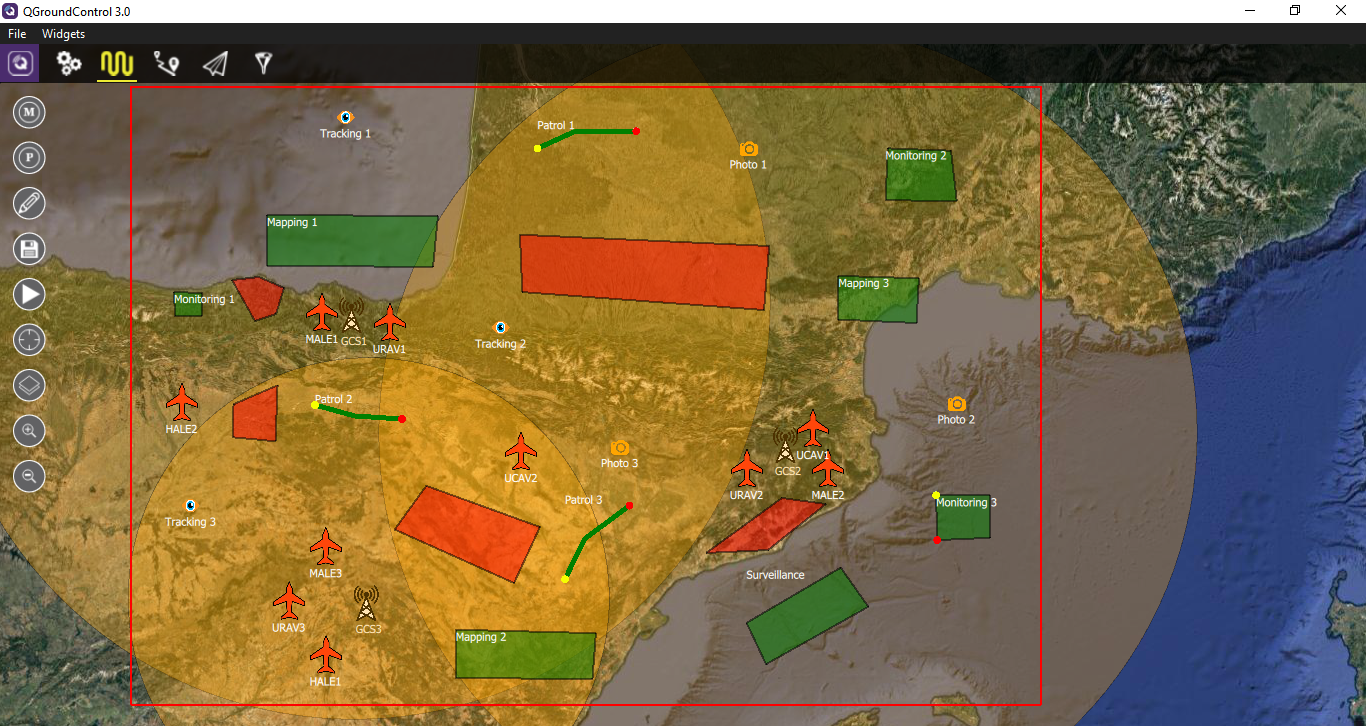}
\caption{Range of GCSs represented as translucent orange circles.}\label{fig:gcs_within_range}
\end{figure}

To facilitate the creation of the mission, \glspl*{gcs} can show a translucent orange circle centred on the station and with radius its within range, graphically representing the range of the \gls*{gcs}. This is shown in Figure \ref{fig:gcs_within_range}.

\subsection{Adding an objective or task}\label{addobjective}
Adding objectives is performed through the third, fourth and fifth buttons in the top of the scenario, depending on whether the location of the objective is a polygon zone, a path or a point, respectively.

When adding a polygon objective, the operator will mark in order the vertices of the zone. Similarly works for a path objective, where the operator marks in order the points of the path. Point objectives, on the other hand, just need to select the single location on the map. Depending on the type of the objective, a different figure will be used to represent the point objective (a fire flame for fire extinguishing, a camera for target photographing, or an eye for target acquisition).

Objectives have different properties including:

\begin{itemize}[leftmargin=*,labelsep=4mm]
\item The name of the objective.
\item The (optional) start and end times of the objective. These are expressed as a date and time, and can only be used when the start time of the mission has been defined.
\item The (optional) duration of the objective. This can only be used with zone and point objectives. When provided, a loiter around the zone or the point will be performed during the specified duration.
\item Whether the objective is mandatory or not.
\item Whether \gls*{los} must be maintained during the objective performance or not.
\item The (optional) entry and exit points of the zone objective (expressed in latitude and longitude). These can only be used in zone objectives.
\item The vertices of the zone or the path (expressed in latitude and longitude).
\item The position of the point (expressed in latitude and longitude). Whether this or the previous property is used, but not both.
\item The type of the objective. This property defines the type of objectives (e.g. target photographing, escorting an individual, fire extinguishing, ...). Depending on this type, a objective may comprise one or more tasks, each one needing a specific sensor for its performance. Depending on the type of task, it must be performed by just one vehicle (e.g. tracking or photographing) or could be performed by several (e.g. mapping or surveillance). In addition, some time or vehicle dependency may be established between the tasks of the objective. These dependencies are discussed in Section \ref{dependencies}.
\end{itemize}

\subsection{Adding a NFZ}\label{addnfz}
A \gls*{nfz} can be added using the sixth button in the top of the scenario. When adding a new \gls*{nfz}, the operator will mark in order the vertices of the zone. This element, and the rest of elements explained in previous sections, can be dragged, and thus their positions and vertices updated. In addition, each vertex can be also dragged. On the other hand, any element of the mission scenario can be deleted by just right clicking over it.

\subsection{Adding objective dependencies}\label{dependencies}
When first entering on edit mode, a general \textit{Mission Info} panel appears on the right, where the initial definition of the mission (bounds, start time, ...) can be modified. In this panel, dependencies between objectives can also be added with the \textit{Add Dependency} button.

A dependency will consist of the following elements:

\begin{itemize}[leftmargin=*,labelsep=4mm]
\item First objective involved in the dependency.
\item The type of dependency. The different types are defined in Allen interval algebra (see Table \ref{tab:allen}).
\item Second objective involved in the dependency.
\item \gls*{uav2} relation between objectives. This value may be undefined, or provided if both objectives must be performed by the same \gls*{uav2} or by different \glspl*{uav2}.
\item Time offset. This property defines the time offset applied between both objectives when considering the dependency (e.g. if an objective precedes another one, then the time offset sets the minimum duration that must pass between the end of the first objective and the start of the second).
\end{itemize}

\begin{table}[!h]
\caption{Allen's interval algebra}
\label{tab:allen}
\centering
\begin{tabular}{|c|c|c|}
 \hline
 Relation    &   Illustration    &   Interpretation
 \\ \hline
 $T_{1} < T_{2}$ & \includegraphics[scale=0.6]{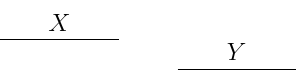} & $T_{1}$ takes place before $T_{2}$
 \\ \hline
 $T_{1}\; m\; T_{2}$ & \includegraphics[scale=0.6]{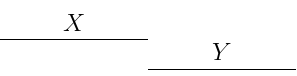} & $T_{1}$ meets $T_{2}$
 \\ \hline
 $T_{1}\; o\; T_{2}$ & \includegraphics[scale=0.6]{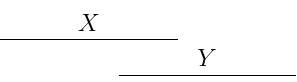} & $T_{1}$ overlaps $T_{2}$
 \\ \hline
 $T_{1}\; s\; T_{2}$ & \includegraphics[scale=0.6]{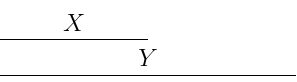} & $T_{1}$ starts $T_{2}$
 \\ \hline
 $T_{1}\; d\; T_{2}$ & \includegraphics[scale=0.6]{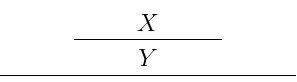} & $T_{1}$ during $T_{2}$
 \\ \hline
 $T_{1}\; f\; T_{2}$ & \includegraphics[scale=0.6]{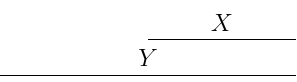} & $T_{1}$ finishes $T_{2}$
 \\ \hline
 $T_{1} = T_{2}$ & \includegraphics[scale=0.6]{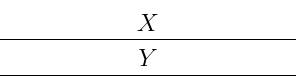} & $T_{1}$ is equal to $T_{2}$
 \\ \hline
 \end{tabular}
\end{table}


\section{Automated Mission Planner and DSS}\label{planner}
The \gls*{mcmpp} consist on assigning each task of the mission the vehicle(s) performing it and the order of performance, and each vehicle the \gls*{gcs} controlling it. In addition, it is also necessary to specify the flight profiles used by the \glspl*{uav2} in each path, and the sensor used in each task, as there could be several sensors on the \gls*{uav2} able to perform the task.

On the other hand, there exist a set of constraints that must be fulfilled to assure the validity of the solutions. These constraints include temporal constraints implying the start and end times of tasks, path constraints assuring that vehicles avoid \glspl*{nfz} in their paths, coverage constraints assuring that the \gls*{uav2} is inside the within range of the \gls*{gcs} controlling it and \gls*{los} is maintained, etc. More information about this problem is presented in \cite{Ramirez2018Weighted}.

In addition, the \gls*{mcmpp} is a \gls*{mop}, as there are multiple variables that must be optimized, including the makespan or end time of the mission, the cost of the mission, the risk, the total fuel consumption of vehicles in the mission, the distance traversed, the flight time, the number of \glspl*{uav2} used and the number of tasks performed. When solving this problem, most of the existing algorithms focus on the approximation of the \gls*{pof}. Nevertheless, when the entire \gls*{pof} comprises a large number of solutions, the process of decision making to select one appropriate solution becomes a difficult task for the \gls*{dmaker}. Sometimes, the \gls*{dmaker} provides a priori information about his/her preferences, which can be used in the optimization process. However, very often the \gls*{dmaker} does not provide this information, and it is necessary to consider other approaches for filtering the number of solutions.

Moreover, a \gls*{dss} is necessary to help the operator in the process of selection of the final plan. This system should provide at least a ranking system and a filtering system. The ranking system considers the operator profile, which provides for each decision variable an intensity factor: \textit{Very low}, \textit{Low}, \textit{Medium}, \textit{High} or \textit{Very high}. Different \gls*{mcdm} methods have been developed over the years for this purpose~\cite{Triantaphyllou2000}.

Once the solutions are ranked, a filtering system based on the distance between the solutions (i.e. the variables of the encoding: the assignments, orders, etc.) is used to erase similar solutions. This distance function must consider the importance of each variable, where assignments are the most important variable, while flight profiles are the least important.

\begin{figure}[H]
\centering
\includegraphics[width=\textwidth]{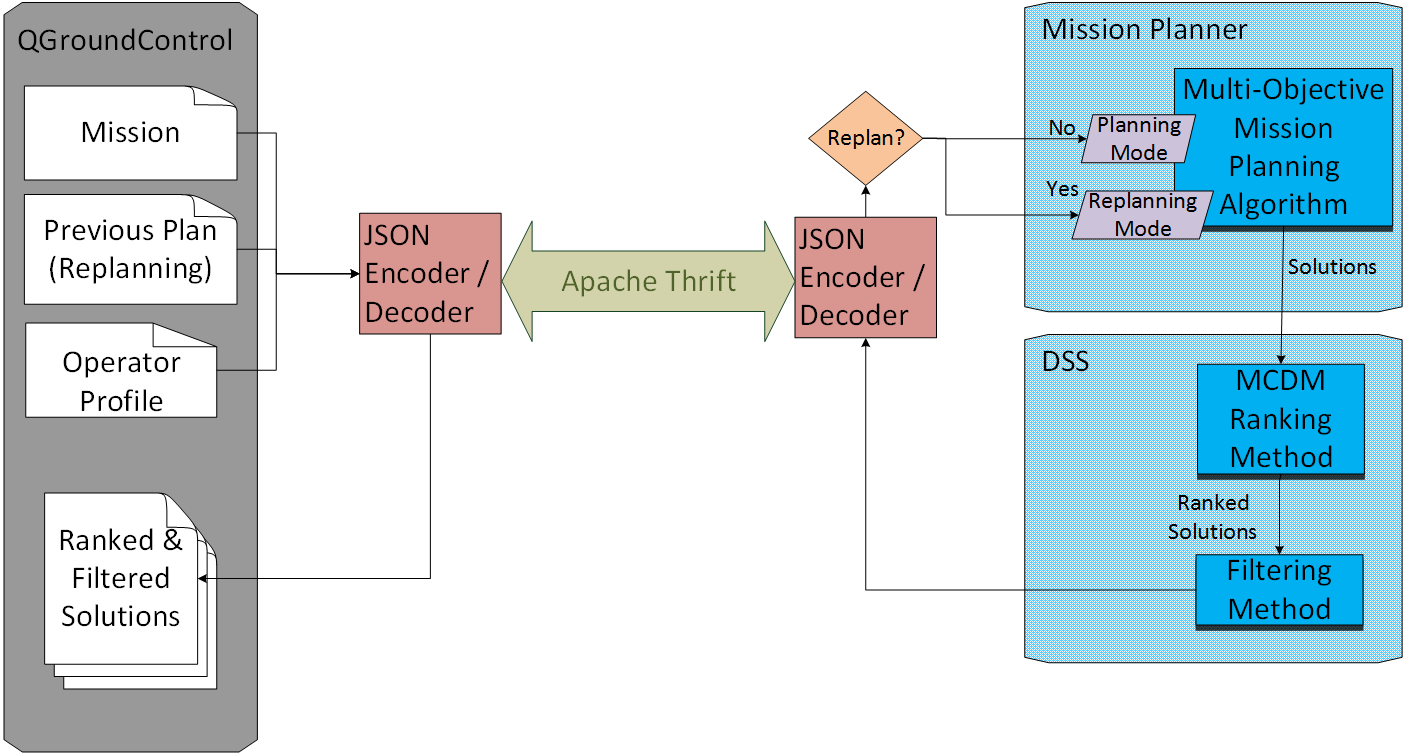}
\caption{Architecture of the Test Bed Interface for Mission Planning and Decision Support.}\label{fig:architectureMP}
\end{figure}

For all this process since the mission is provided until the ranked and filtered solutions are returned, a test bed interface has been designed. The communication interface between QGroundControl and the planning and decision algorithms has been implemented using Apache Thrift\footnote{https://thrift.apache.org/}. This frameworks permits an easy communication with most known programming languages. The interface sends the mission and the operator profile extracted from QGroundControl as a JSON message. This message includes the different parameters of each element of the mission explained in Section \ref{designer}. On the other hand, the message returned by the algorithm must contain a ranked list of solutions, where each solution defines the assignments for each task, including the flight profile and sensors used; the \gls*{gcs} assignment, final path and performance variables (fuel consumption, flight time, etc.) for each \gls*{uav2}, and the values of the optimization objectives and risk factors of the problem. The architecture of these modules is represented in Figure \ref{fig:architectureMP}.

\subsection{Operator Profile}\label{user_profile}
Every operator using the QGroundControl has a profile defining the different constraints, fitness and ranking variables predefined. This profile includes the following settings:

\begin{itemize}[leftmargin=*,labelsep=4mm]
\item Whether all tasks must be performed or not.
\item Minimum and risked distance to ground: These variables define the interval for the risk factor distance to ground, where the minimum represented a risk of 100\% and values higher than the risked represent a 0\% risk.
\item Maximum and risked percentage of fuel usage: These variables define the interval for the risk factor fuel usage per \gls*{uav2}, where the maximum represented a risk of 100\% and values lower than the risked represent a 0\% risk.
\item Minimum and risked distance between vehicles: These variables define the interval for the risk factor distance between \glspl*{uav2}, where the minimum represented a risk of 100\% and values higher than the risked represent a 0\% risk.
\item Minimum and maximum risked time out of \gls*{gcs} coverage: These variables define the interval for the risk factor time out of \gls*{gcs} coverage, where the minimum represented a risk of 0\% and the maximum risked represent a 100\% risk.
\end{itemize}

On the other hand, the operator must also define the importance (very low, low, medium, high or very high) of the ranking variables to be used by the \gls*{dss}. These variables include:

\begin{itemize}[leftmargin=*,labelsep=4mm]
\item Makespan or end time of the mission.
\item Total cost of the mission.
\item Total fuel consumption of vehicles in the mission.
\item Total flight time of the vehicles in the mission.
\item Total distance traversed by the vehicles in the mission.
\item Risk of high fuel usage. This considers the \glspl*{uav2} that finish the mission with low fuel.
\item Risk of low distance to ground. This considers the vehicles that fly near to the ground (depending on the route and the altitude of the adopted flight profile).
\item Risk of \gls*{gcs} coverage loss. This considers \glspl*{uav2} that fly out of the coverage or \gls*{los} of the \glspl*{gcs} controlling them
\item Risk of \gls*{uav2} closeness. This considers vehicles that fly close between them, which intuitively depends on the time constraints between concurrently performed tasks and eventual spatial overlaps among routes/flight profiles.
\item Number of \glspl*{uav2} employed in the mission.
\item Number of tasks performed. This is considered when some tasks or objectives are not mandatory.
\item Number of \glspl*{gcs} employed in the mission.
\end{itemize}

\subsection{Mission Planning}
Once a mission is defined, the mission planner (button ``P'' on the left panel in Figure \ref{fig:mission}) can be executed in order to find plans for this mission. Additionally, the operator preferences defined in Section \ref{user_profile} can be adapted for this concrete mission. Apart from the preferences explained before, the operator can also specify some constraints for the mission:

\begin{itemize}[leftmargin=*,labelsep=4mm]
\item Maximum makespan: The maximum valid makespan. Plans with higher makespan will be rejected from the solutions.
\item Maximum cost: The maximum valid cost. Plans with higher cost will be rejected from the solutions.
\item Maximum flight time: The maximum valid flight time. Plans with higher flight time will be rejected from the solutions.
\item Maximum fuel consumption: The maximum valid fuel consumption. Plans with higher fuel consumption will be rejected from the solutions.
\item Maximum distance traversed: The maximum valid distance traversed. Plans with higher distance traversed will be rejected from the solutions.
\end{itemize}

Once this information is completed, it is encoded as a JSON message, and sent through the Thrift interface to the automated mission planner, which must detect it as preplanning and deal with multiple objectives.

\subsection{DSS ranking and filtering}
After the mission planner finishes, the \gls*{dss} will take the solutions obtained and rank them according to the ranking criteria defined by the operator using some \gls*{mcdm} method. Then, the \gls*{dss} will filter the solutions that are very similar (only differ in the flight profile used in some path, the sensor employed, ....).

\begin{figure}[H]
\centering
\includegraphics[width=0.9\textwidth]{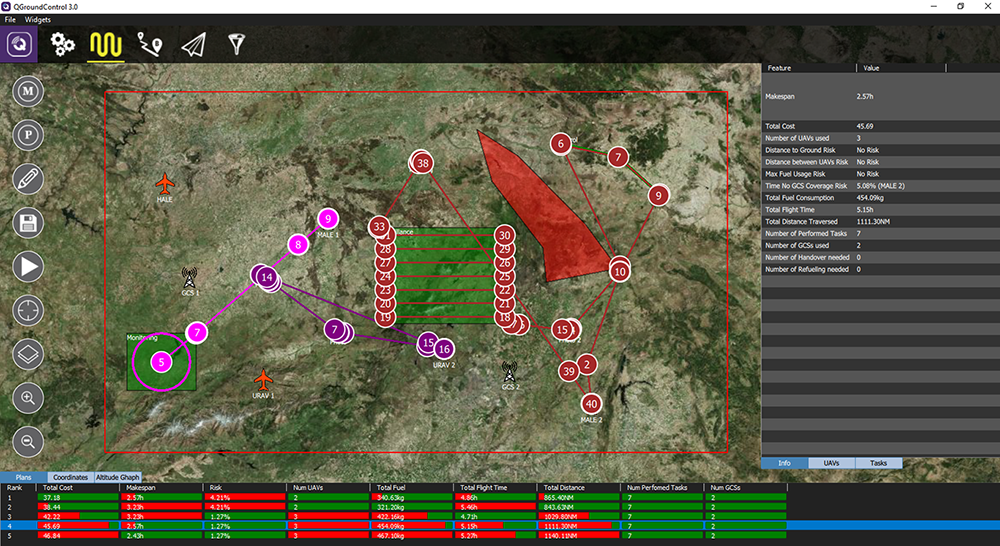}
\caption{Mission Plans.}\label{fig:plans}
\end{figure}

After this, the new set of ranked-filtered solutions will be returned as a JSON message to QGroundControl, which decodes this message and presents the set of plans in a table in the bottom of the scenario (see Figure \ref{fig:plans}). This table shows the different objectives optimized in a percentage bar, where more green bars represent better values for the variable while more red bars represent worse values for the variable.

\subsection{Plan Visualization}\label{visualization}
Once the plans have been computed, it is possible to view the paths and some information about a specific plan by clicking on it. The path for each \gls*{uav2} used in the mission is represented with a different color (see Figure \ref{fig:plans}). On the other hand, the right pop-up shows three tabs with different information about the plan. The info tab shows the different values for the objective variables and risk factors of the mission plan.

The \textit{UAVs} tab shows for every \gls*{uav2} used in the mission (represented in columns), information about the \gls*{uav2} assignments and performance:

\begin{itemize}[leftmargin=*,labelsep=4mm]
\item The assigned tasks (in order) that the \gls*{uav2} performs.
\item The GCS assigned to the \gls*{uav2} in the mission.
\item The departure time for the \gls*{uav2} in the mission.
\item The return time for the \gls*{uav2} in the mission, and the return flight profile (FP) used.
\item The cost of use of the \gls*{uav2} in the mission.
\item The flight time of the \gls*{uav2} in the mission.
\item The distance traversed by the \gls*{uav2} in the mission.
\item The fuel consumed by the \gls*{uav2} in the mission.
\item The percentage of fuel usage of the \gls*{uav2}.
\item The minimum distance to the ground of the \gls*{uav2} during the mission.
\item The time spent by the \gls*{uav2} out of the \gls*{gcs} coverage during the mission.
\end{itemize}

On the other hand, if clicking on the \textit{Tasks} tab, a table will appear showing the task assignments in files, where each file provides:

\begin{itemize}[leftmargin=*,labelsep=4mm]
\item The objective considered
\item The task of the objective considered (at least one)
\item The \gls*{uav2}(s) assigned to the task.
\item The departure time for the \gls*{uav2} when it start going to the task zone.
\item The flight profile used in the path to reach the task zone.
\item The wait loiter duration in case the task has some time restriction and the vehicle must wait until its start.
\item The start time of the task.
\item The sensor used for the performance of the task.
\item The duration of the task.
\item The end time of the task.
\end{itemize}

By left clicking one of the \glspl*{uav2} used in the mission plan, only the path for this \gls*{uav2} will be represented. In addition, if there exist some out of coverage points for the route of the \gls*{uav2}, these will be represented in red. The tables of the right panel will also adapt to this selection and only show the concrete \gls*{uav2} and its tasks assigned.

When clicking the \textit{Coordinates} tab in the bottom, a table with the different waypoints of the route of the selected \gls*{uav2} will show every specific parameter of each waypoint, such as the speed, time, the task associated to it, etc.

On the other hand, when clicking the \textit{Altitude Graph} tab on the bottom, an altitude profile for the selected \gls*{uav2} will appear, including the ground altitude.


\section{Mission Execution and Replanning}\label{replanning}
After visualizing the ranked plans, the operator selects the best one according to its criteria (should be usually the first one) and this plan will be simulated by clicking the play button on the left panel (see Figure \ref{fig:mission}).

After this, several instances of the ArduPilot program will be executed (as many as \glspl*{uav2} used in the mission). Each instance will be connected with the QGroundControl through a MAVLink connection, and the different \gls*{uav2} figures will be associated with the position of the related ArduPilots. So the \glspl*{uav2} will start departing according to the plan.

As the waypoints in the path for each vehicle are passed by, they will turn into a darker color, and their border will become black. The current waypoint where the \gls*{uav2} is going is highlighted in green. On the other hand, the tasks that finish their performance become more translucent.

During the execution of the mission, if the operator receives any external notification about an event involving a new objective, he can enter in edit mode and add new objectives as explained in Section \ref{designer}.

Once the new tasks are added, the mission planner can be executed similarly as in preplanning. As a previous plan is being simulated, the JSON message sent to the mission planner must contain this previous plan, so the planner knows that it must work in replanning mode. In addition, the operator introduces a concrete time for the planning process. This time will not only limit the planning process, but also will be considered as the moment where the replanning process is performed (i.e. the status of the mission passed to the replanner will be the one taking place those seconds after the actual moment).

Then, as in the planning process explained in previous section, the \gls*{dss} ranks and filters the solutions and these are returned to the operator, where the different paths and assignments can be seen as before.

Finally, once the operator selects the new plan to be updated, the paths and assignments for the actual executing mission plan will be updated.


\section{Use cases on the extended QGroundControl}\label{casestudy}
In this section, we design two use cases to prove the new functionalities added to QGroundcontrol. In the first one, a general walk through of the different tools developed is done, creating a mission, planning, simulating and replanning it. In the second one, a mission that is impossible to solve is presented, in order to show how the planner informs the operator in this situation.

In order to use the test bed interface, a novel mission planning algorithm~\cite{Ramirez-Atencia2017Knee} has been used. This algorithm extends \gls*{nsga2}~\cite{Deb2002Fast} to focus the search on "knee points" \cite{Branke2004Finding}, thereby looking for the most significant solutions in the \gls*{pof}. This approach checks the validity of solutions through a \gls*{csp} model developed using Gecode \cite{Schulte2010Modeling}, which is connected with the fitness function of the algorithm. Moreover, the replanning algorithm used~\cite{Ramirez-Atencia2016MOGAMR} is the same approach as mission planning but taking into account the previous plan and the limited time for the algorithm.

On the other hand, for the \gls*{dss}, VIKOR \cite{Opricovic2004Compromise} has been used to rank the solutions returned by the \gls*{moea}, using the factors defined by the operator profile as the weights of the criteria. The VIKOR method uses the Manhattan distance and the Chebyshev distance, and provides a compromise solution, considering the maximum utility and the minimum individual regret. Finally, the filtering is performed through a distance function that assigns a weight to each variable based on its importance. When two solutions are separated less than a filter threshold, the one with the lower rank value is omitted. The resultant
set of ranked and filtered solutions is then returned through the Thrift interface.

\subsection{First use case: A walking through the framework}
First, to create a new mission, we use ``M'' button in the left panel to access the read or create mission panel (see Figure \ref{fig:create_read_mission}). Here, we must provide the parameters mentioned in Section \ref{create}. In this case, the bounds are latitude between $36.76^\circ$ and $36.85^\circ$ and longitude between $-2.396^\circ$ and $-2.174^\circ$, and 7.5 arc-seconds are used in the elevation map. The start time of the mission is not specified. Then by clicking the Create Mission button (it will be available as long as all the mandatory values are fulfilled and correct), the mission will be created and an empty scenario inside the defined bounds will be presented.

Now, to recreate the mission presented in Figure \ref{fig:mission}, it is necessary to add the objectives, \glspl*{uav2}, \glspl*{gcs} and \glspl*{nfz}. These instances are added using the Mission Designer, selecting the Edit button represented with a pencil in the left panel. Then, the set of icons in the top of the scenario is used to add each of the elements of the mission. In this case, we consider 4 \glspl*{uav2}, 1 \gls*{gcs}, 1 \gls*{nfz} and 5 objectives (monitoring, surveillance, patrol, tracking and target photographing). To add the vehicles, just clicking on the corresponding icon on the top and then on the desired position in the scenario, the elements will be positioned and a pop-up will appear to specify the mandatory properties of these elements (e.g. Figure \ref{fig:add_uav} shows this pop-up for a \gls*{uav2}, where a name for this vehicle must be provided, as well as its type and configuration). For the \glspl*{gcs} and the objectives, this pop-up will require a name and the type of the station or the objective, while \glspl*{nfz} do not require any parameter, so no pop-up will appear when creating them. As was mentioned in \ref{addobjective}, when adding zone and path objectives, and also when adding \glspl*{nfz}, after clicking the corresponding icon for these elements, a set of points in the scenario must be clicked in order to create the desired zone or path.

If the elements created are not properly positioned, all of them can be dragged, and thus their positions and vertices updated. In addition, each vertex for path and zone objectives can be moved using the orange points that appear at each vertex in edit mode. The entry point (yellow) and exit point (red) for zone objectives can also be dragged. On the other hand, any element of the scenario can be deleted by just right clicking over it.

Once the elements have been created, to modify their different properties, just left clicking on them will trigger a panel in the right, showing the properties of the element (e.g. Figure \ref{fig:edit_zone} shows the properties of the monitoring objective, where entry and exit points have been added and a duration of 15 minutes has been established). Once these properties are modified, the Save button must be clicked and a message ``Saved successfully'' must appear, or an error message indicating the possible error. The tracking objective has a duration of 10 minutes, and the rest of the objectives have not been modified. On the other hand, the \glspl*{uav2} considered are a HALE (with \gls*{eoir} camera and \gls*{sar} radar) two URAV (with \gls*{eoir} camera) and a MAlE (with \gls*{mpr} and \gls*{sar} radars).

When first entering on edit mode, a general \textit{Mission Info} panel appears on the right, where the initial definition of the mission (bounds, start time, ...) can be modified. In this panel, dependencies between objectives can also be added by clicking the \textit{Add Dependency} button, and deleted with the ``X'' button above the concrete dependency. As can be seen in Figure \ref{fig:mission}, we consider two dependencies on this mission, establishing that the surveillance objective (Search Area) must precede the target photographing and tracking objectives, and the tracking objective must be performed by the same \gls*{uav2} that performs the surveillance. To save the new dependencies added or deleted, the Save button of the panel must be clicked.

Once the entire mission has been defined, in order to save it on the database, the \textit{Save} button on the left (the one with the floppy disk) is used.

Once the mission is defined, the mission planner can be executed in order to find solutions for this mission. This is done by clicking the ``P'' button on the left (see Figure \ref{fig:plan_mission}). Then, a pop-up will appear showing previous executions of the mission planner in the top, and a button for executing the planner (including a box for indicating the maximum runtime) in the bottom.

Additionally, the Change Config button permits to change the operator preferences defined in Section \ref{user_profile} for this concrete mission (see Figure \ref{fig:change_config}). In this case, the ranking values are assigned a very high value for the makespan; high values for cost, fuel consumption, flight time, distance and number of tasks performed; medium value for the number of vehicles used, and low value for the risk factors and the number of stations used. When clicking the \textit{Plan Mission} button, the automated mission planner starts running with the mission designed. 

After the mission planner gets the solutions, and the \gls*{dss} ranks and filters them, they are presented in a table in the bottom of the scenario (see Figure \ref{fig:plan_visualization}). Each row of the table represent a solution, and each column represent the value of an optimization variable. The cells are filled as a percentage bar, where the greener a cell is, the better the optimization variable for that solution with respect to the others. When clicking one of the rows of this table, the paths for each vehicle will be shown, and a panel in the right will appear presenting some information about the plan, including the risk factors. This panel has two extra tabs: the \textit{UAVs} tab presents information about the performance of every specific \gls*{uav2}, including the departure time of the vehicle, the fuel consumed, the assigned tasks, etc; while the \textit{Tasks} tab presents information about every task, including the vehicles performing it, the flight profile used by them, the sensor employed, etc.

When selecting one of the \glspl*{uav2} used in the plan, only the path for this \gls*{uav2} will be represented (see Figure \ref{fig:plan_coordinates}). In this situation, the \textit{Coordinates} tab in the bottom will present a table with the different waypoints of the selected vehicle, including parameters such as the speed, estimated time of arrival, etc.

On the other hand, when clicking the \textit{Altitude Graph} tab on the bottom (see Figure \ref{fig:plan_altitude}), an altitude profile for the selected \gls*{uav2} (in yellow) will appear, including the ground altitude (in brown).

Now that the returned plans have been studied, we select the best one, which usually should be the first one as in this case. Then, to simulate this plan, we click the play button on the left panel (see Figure \ref{fig:mission_execution}). Then a pop-up indicates that the ArduPilots simulating the vehicles are being initialized and the paths are being loaded to them.

Once the simulation start, it can be seen how the vehicles in the scenario start moving and the waypoints where they are going are marked in green. Once these waypoints are traversed, their color becomes darker and their border turns black. Meanwhile, when tasks are completed, they become more translucent (see Figure \ref{fig:online_edit}). When the exact moment comes, in our case it is represent Figure \ref{fig:online_edit}, we simulate that two new tasks must be performed as soon as possible, so we add them to our current mission. To do that, we select the Edit button and add new objectives. In this case, an Oil Leaks monitoring and a New Photo objectives have been added.

After the online editing, we click the Planner button and introduce a concrete time for the Mission Planning process (usually 1 or 2 minutes). Then, by clicking the Plan Mission button, the mission replanner will be executed during the specified time. When this process finishes, new solutions will be returned in the form of a table (see Figure \ref{fig:mission_replanned}), where the different paths and assignments can be seen as before.

In this case, the optimization process has returned one solution. This new plan involves a new vehicle to perform the Oil Leaks Monitoring objective and the re-routing of URAV 2 for performing the New Photo objective. Now, we click the \textit{Re-Execute} button (see Figure \ref{fig:mission_reexecution}), and the paths and assignments for the actual executing mission plan will be updated.

\subsection{Second use case: Working with unresolvable missions}
In this case, we consider a new mission, represented in Figure \ref{fig:case2}, with 3 \glspl*{uav2}, 1 \glspl*{gcs}, 1 \gls*{nfz} and 6 objectives. As can be seen, the mission cannot be performed because the within range of the \gls*{gcs} does not cover all the objectives of the mission.

In this case, when performing the planning process, following the same steps as in the previous case, the mission planner will not return any solution. Instead, a pop-up will appear (see Figure \ref{fig:case2_no_solutions}), informing that no solution was found, and the errors that occurred most frequently during the checking in the \gls*{csp} model, so they represent the most probable problem that presents this mission, so the operator can reformulate the mission to make it feasible. As can be seen in Figure \ref{fig:case2_no_solutions}, the planner informs that the main problem is that the vehicles spent too much time out of the coverage of the \gls*{gcs}, as was pointed before. Also, it is appreciable that the fuel of the vehicles may be insufficient in some plans.


\section{Conclusions}\label{conclusions}
In this work, we have extended the QGroundControl framework, adding the functionalities defined in Figure \ref{fig:architecture}. The contribution of this work consists of:

\begin{enumerate}[leftmargin=*,labelsep=3mm]
\item Design and development of a Mission Designer, which provides an interactive environment for the creation and visualization of missions, including its objectives/tasks, vehicles, \glspl*{gcs}, \glspl*{nfz}, etc. 
\item Integration of an interface for automated mission planner and \gls*{dss}, in order to test different mission planning and \gls*{dss} algorithms, which generate, rank and filter plans for the missions designed.
\item Design and development of a plan visualizer, which permits to graphically represent the plans, including the paths for each \gls*{uav2} and the information related to the optimality and risks of the plan.
\item Design and development of a mission monitoring system, which informs the operator about the waypoints already passed by and the tasks already performed.
\item Design and development of a replanning system and integration of the automated mission replanner inside the QGroundControl by reusing the interface for automated mission planning. This permits the operator to inform the system about new objectives/tasks or incidents during the execution of the mission and call the mission replanner in order to obtain new plans for the updated mission.
\end{enumerate}

In future works, this framework will be outperformed adding other novel techniques that are being developed for \glspl*{uav2}, such as a training system for operators, the use of new controlling devices (e.g. virtual reality glasses or motion sensing devices) or inclusion of augmented reality in the simulation.

\vspace{6pt} 


\acknowledgments{This work has been supported by the next research projects: EphemeCH (TIN2014-56494-C4-4-P) and DeepBio (TIN2017-85727-C4-3-P) by Spanish Ministry of Economy and Competitivity (MINECO), both under the European Regional Development Fund FEDER, and by Airbus Defence \& Space (FUAM-076914 and FUAM-076915). The authors would like to acknowledge the support obtained from Airbus Defence \& Space, specially from Savier Open Innovation project members: Jos\'e Insenser, Gemma Blasco and C\'esar Castro.}

\authorcontributions{Investigation, C.R.A.; Writing-original draft, C.R.A.; Supervision, D.C.}

\conflictofinterests{The authors declare no conflict of interest.} 



\appendix
\section{Use cases (Screenshots)}
Here we provide the different screenshots from the simulator execution, used in the use cases presented in Section \ref{casestudy}.

\begin{figure}[H]
\centering
\includegraphics[width=\textwidth]{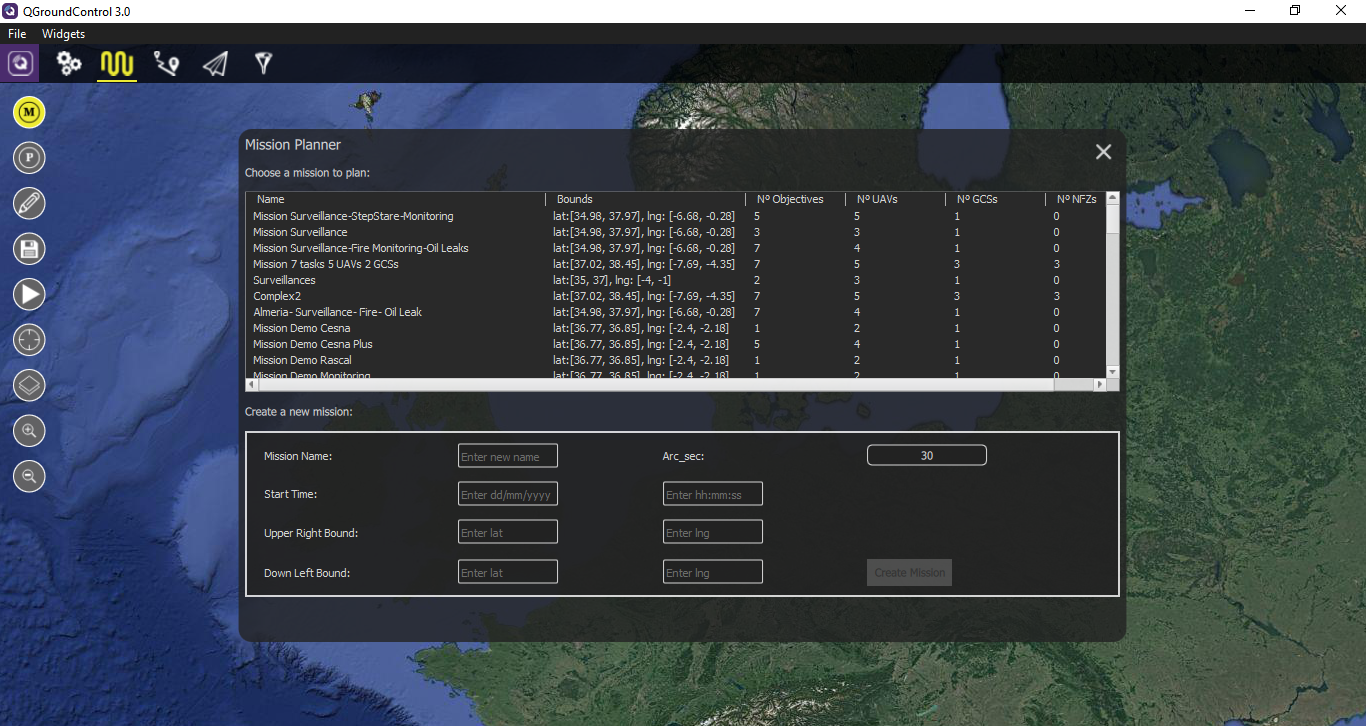}
\caption{Panel in the QGroundControl for reading or creating missions.}\label{fig:create_read_mission}
\end{figure} 

\begin{figure}[H]
\centering
\includegraphics[width=\textwidth]{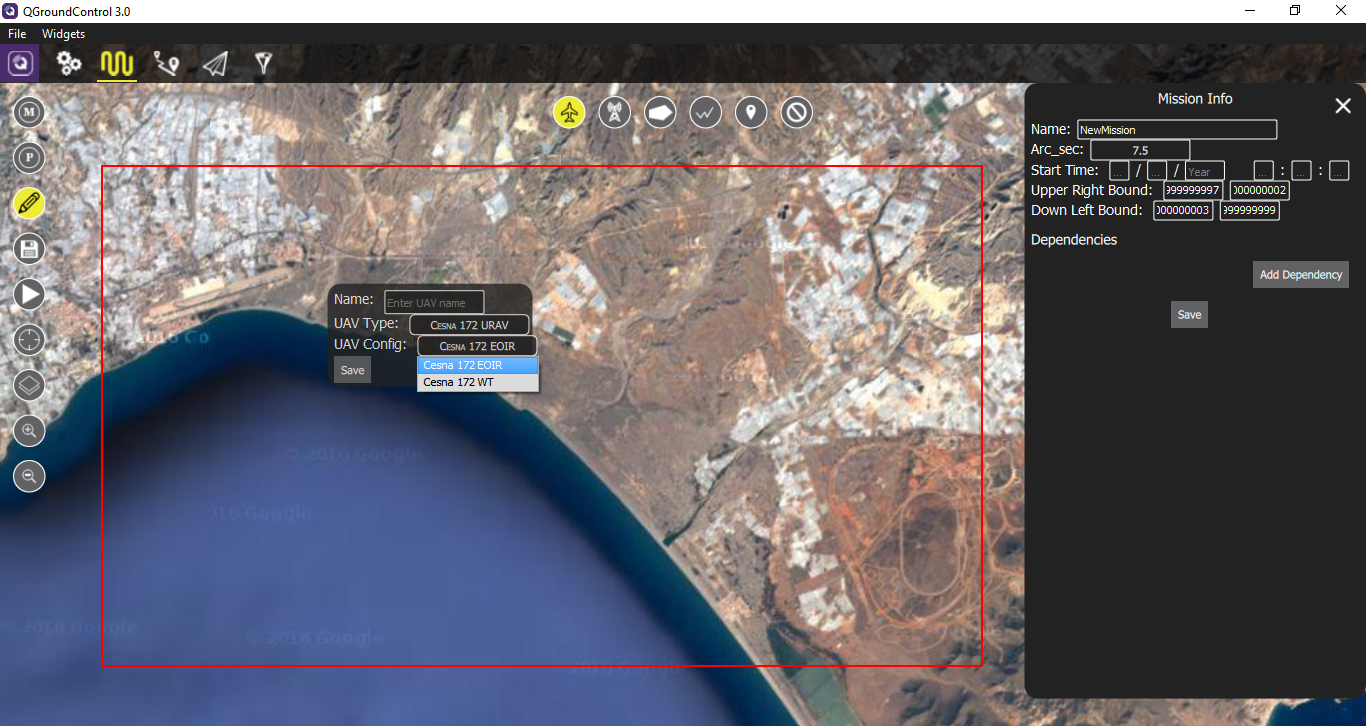}
\caption{Adding a new UAV in the Mission Designer.}\label{fig:add_uav}
\end{figure} 

\begin{figure}[H]
\centering
\includegraphics[width=\textwidth]{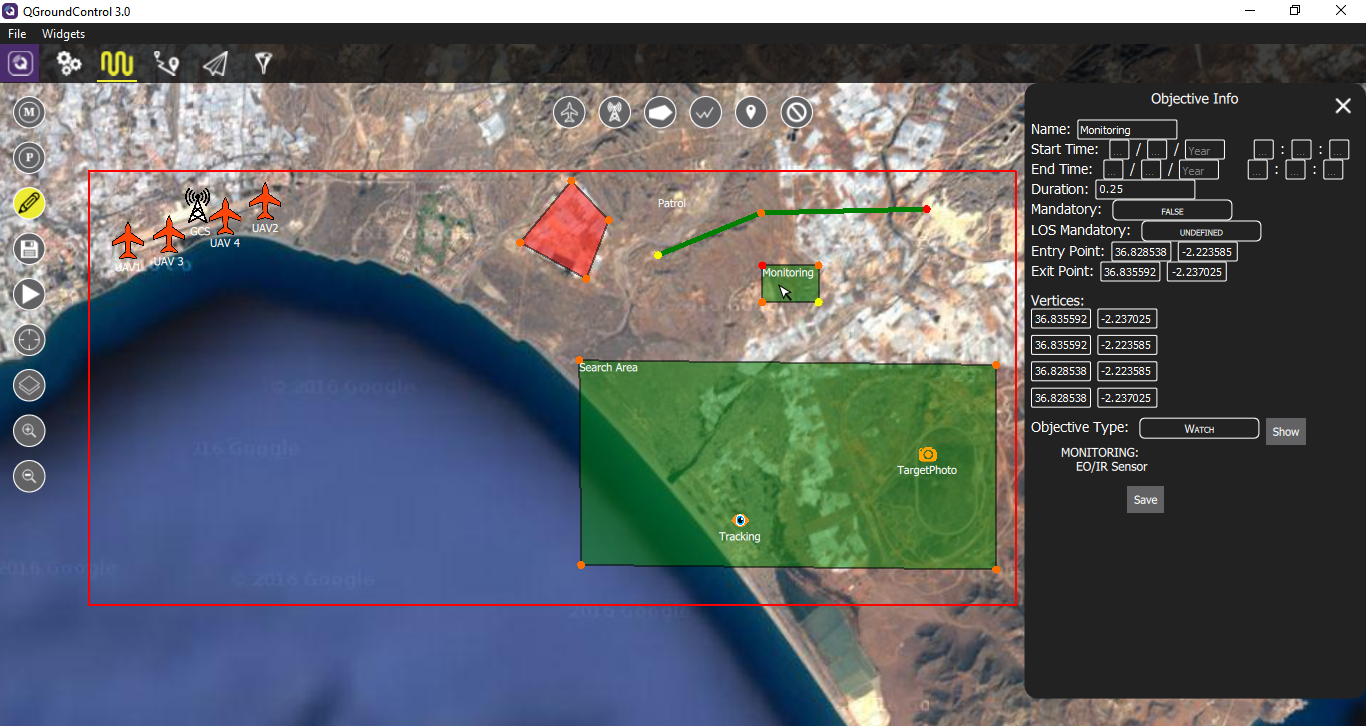}
\caption{Modifying zone objective properties in the Mission Designer.}\label{fig:edit_zone}
\end{figure} 

\begin{figure}[H]
\centering
\includegraphics[width=\textwidth]{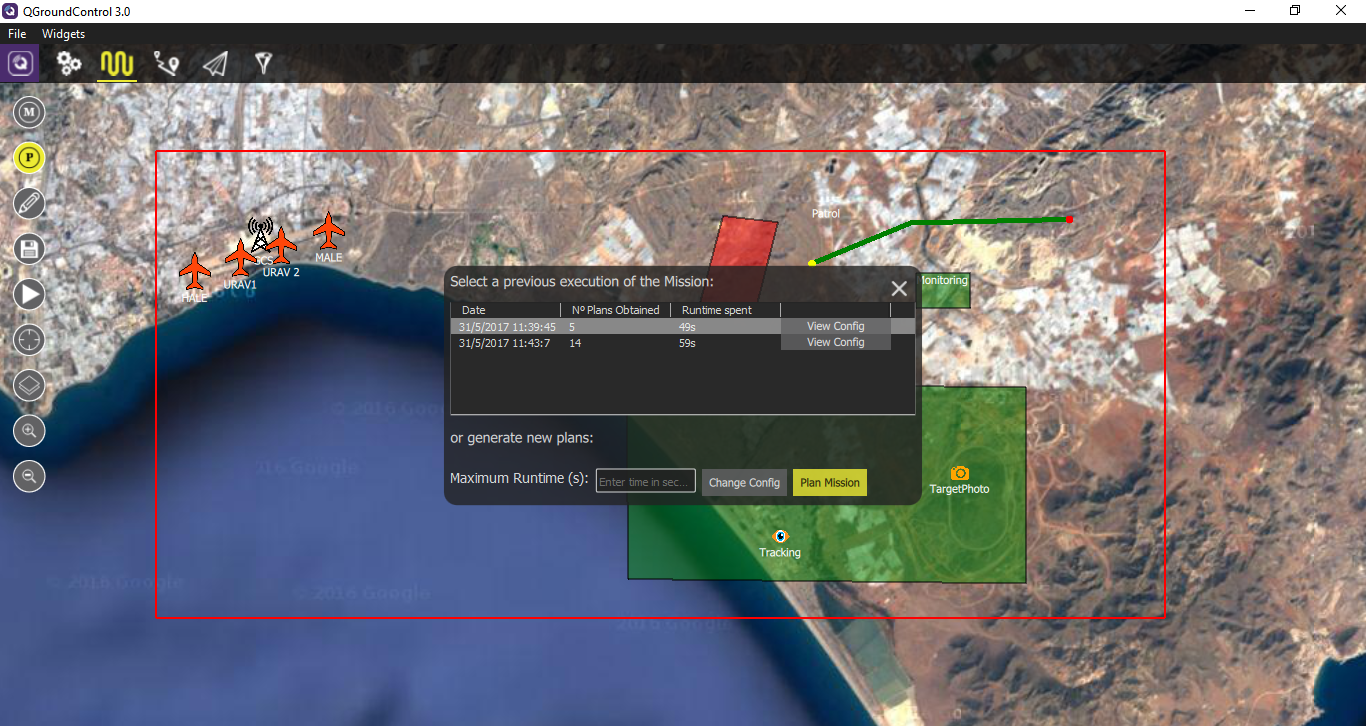}
\caption{Using the Automated Mission Planner.}\label{fig:plan_mission}
\end{figure} 

\begin{figure}[H]
\centering
\includegraphics[width=\textwidth]{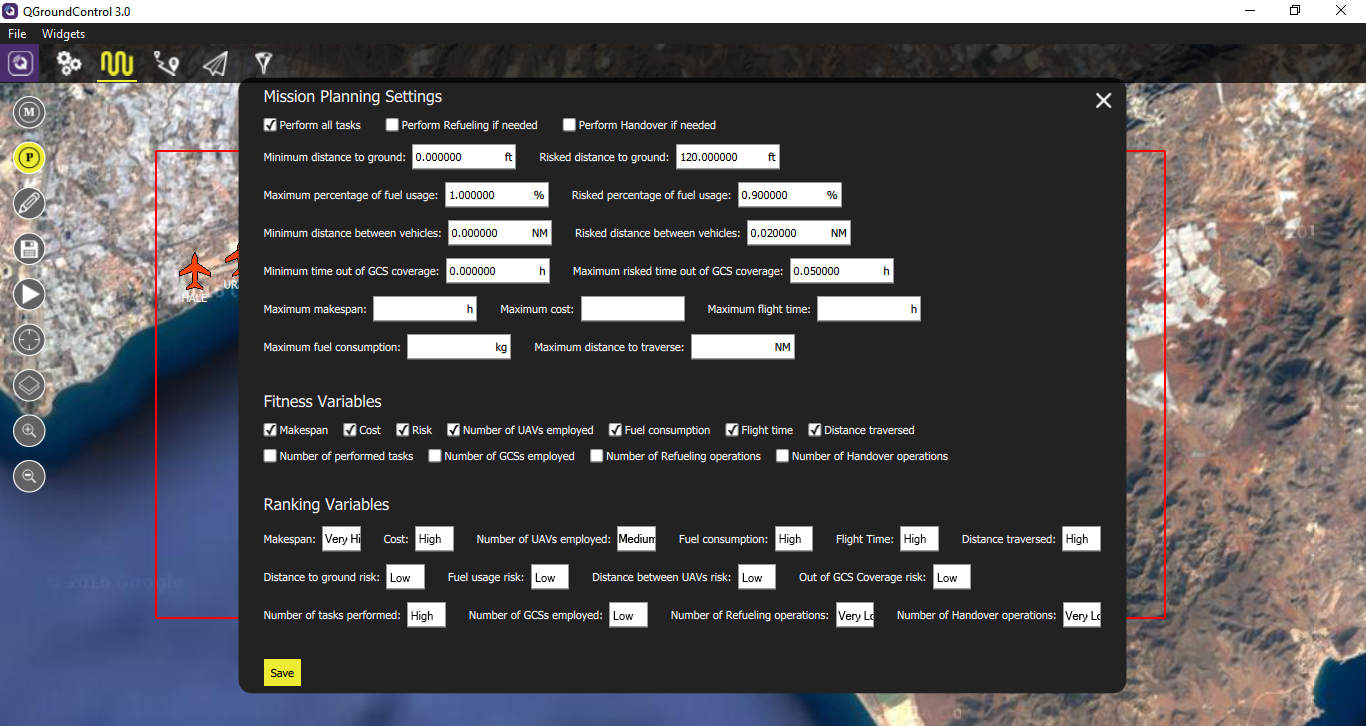}
\caption{Changing the Operator Settings for the mission.}\label{fig:change_config}
\end{figure}

\begin{figure}[H]
\centering
\includegraphics[width=\textwidth]{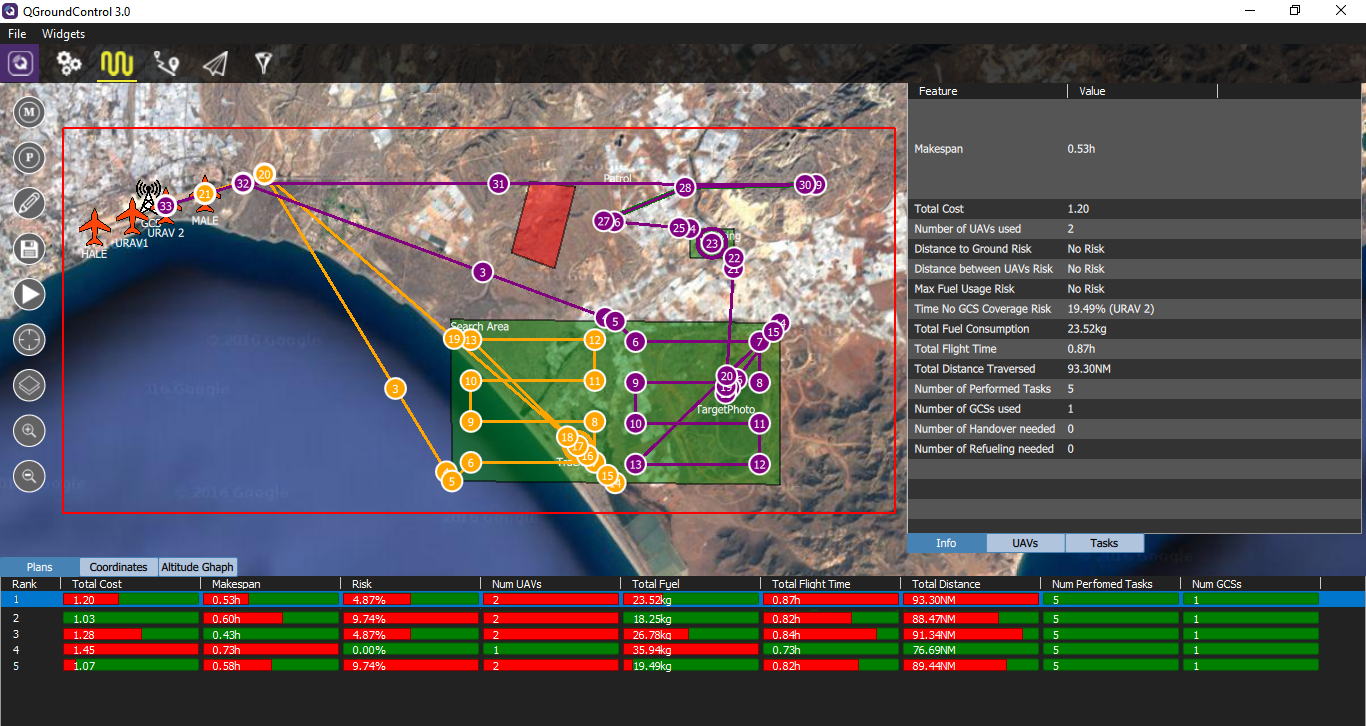}
\caption{Mission Plan routes and info.}\label{fig:plan_visualization}
\end{figure} 

\begin{figure}[H]
\centering
\includegraphics[width=\textwidth]{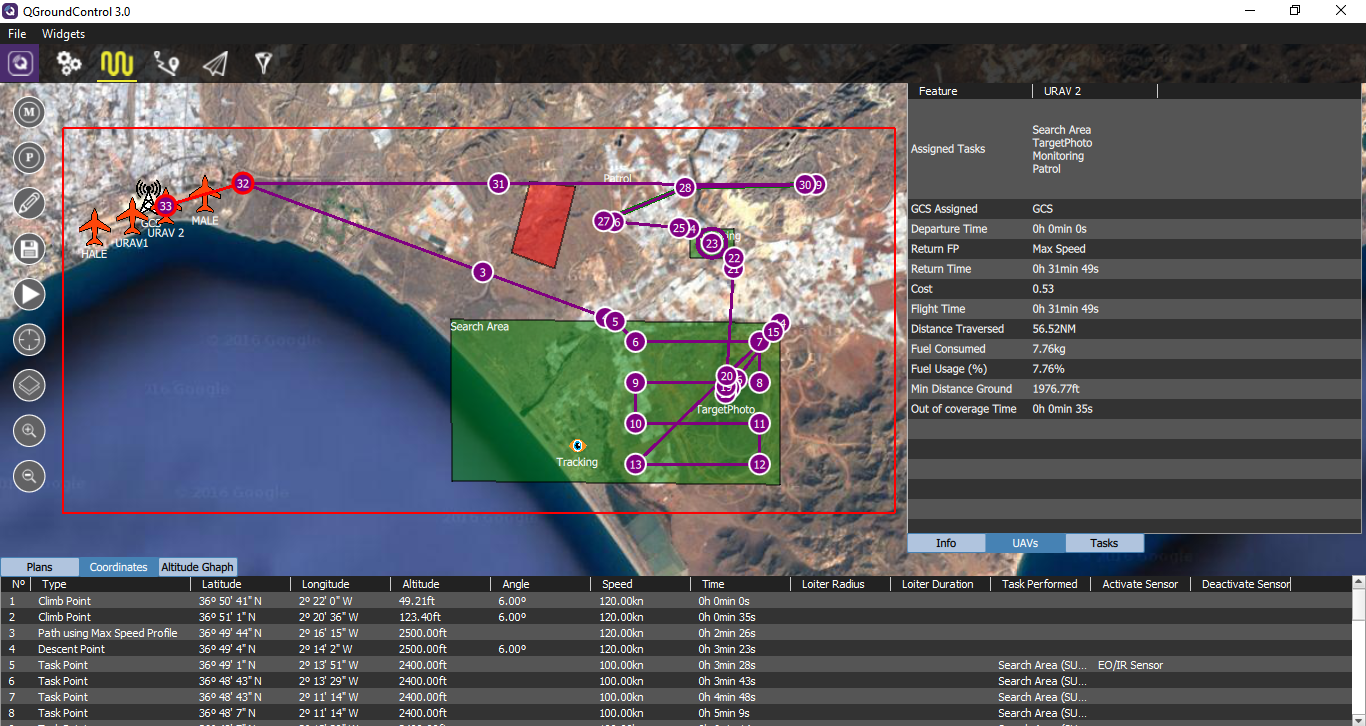}
\caption{Route for URAV 2, including waypoint table and UAVs info.}\label{fig:plan_coordinates}
\end{figure} 

\begin{figure}[H]
\centering
\includegraphics[width=\textwidth]{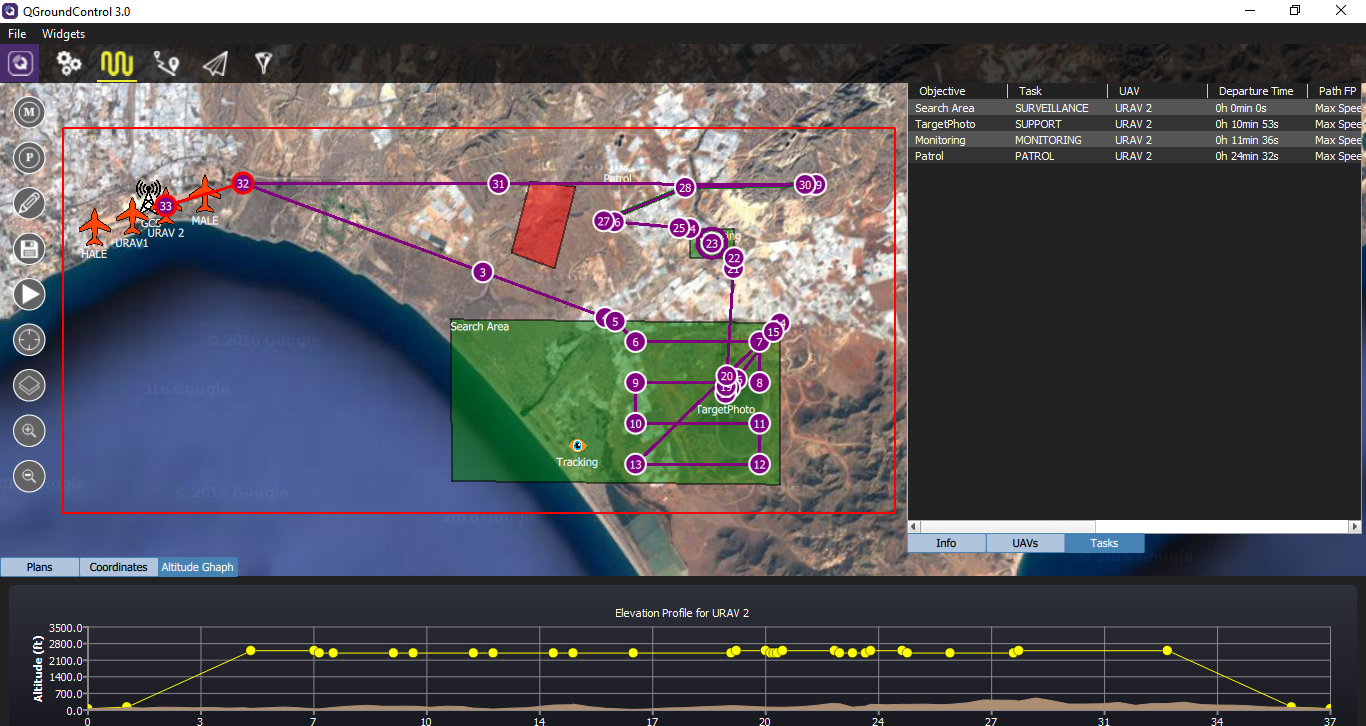}
\caption{Altitude Profile for URAV 2 and tasks table.}\label{fig:plan_altitude}
\end{figure} 

\begin{figure}[H]
\centering
\includegraphics[width=\textwidth]{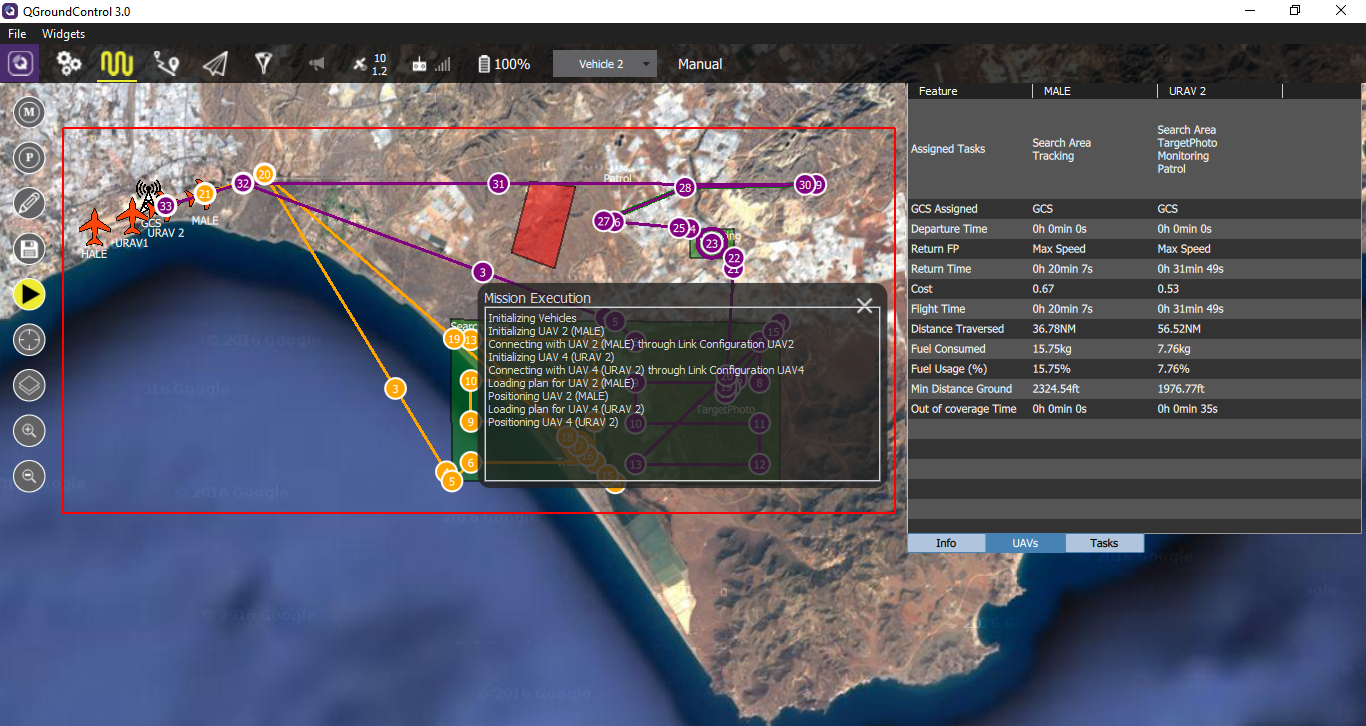}
\caption{Mission Execution.}\label{fig:mission_execution}
\end{figure} 

\begin{figure}[H]
\centering
\includegraphics[width=\textwidth]{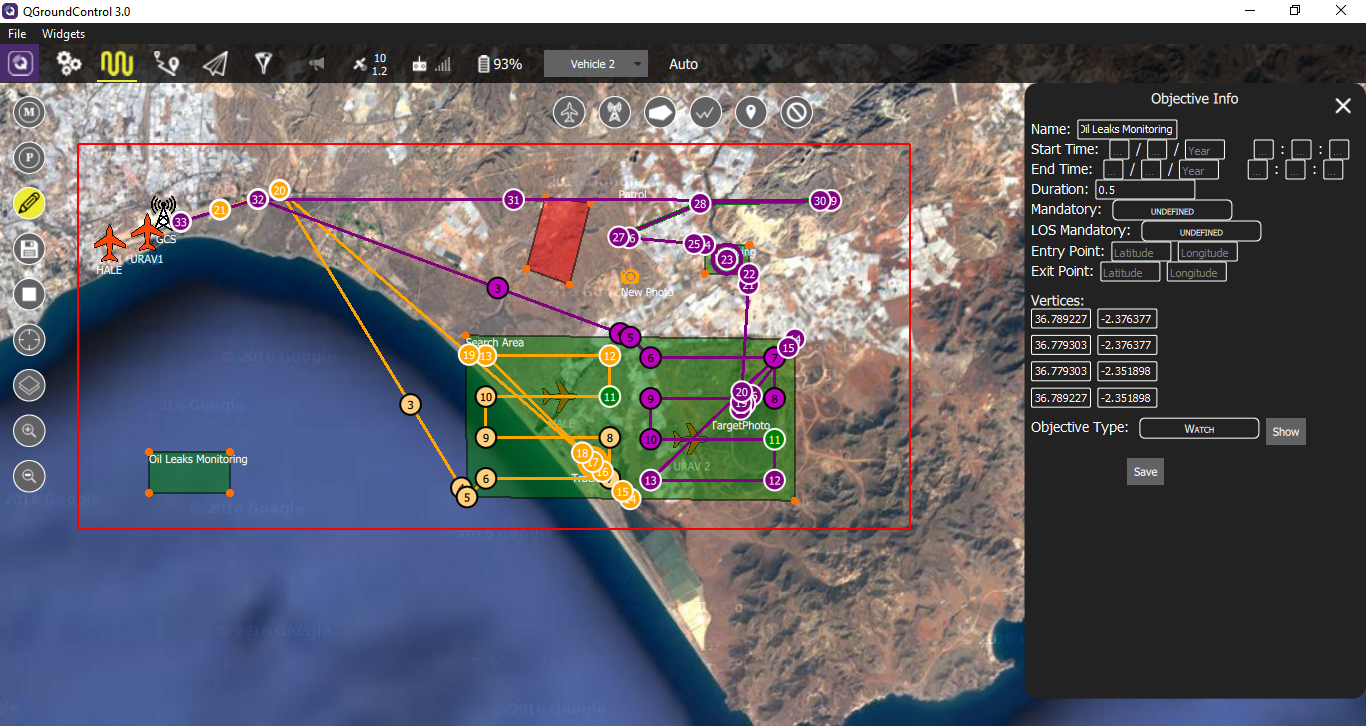}
\caption{Mission Online Edition. Adding new objectives during execution.}\label{fig:online_edit}
\end{figure} 

\begin{figure}[H]
\centering
\includegraphics[width=\textwidth]{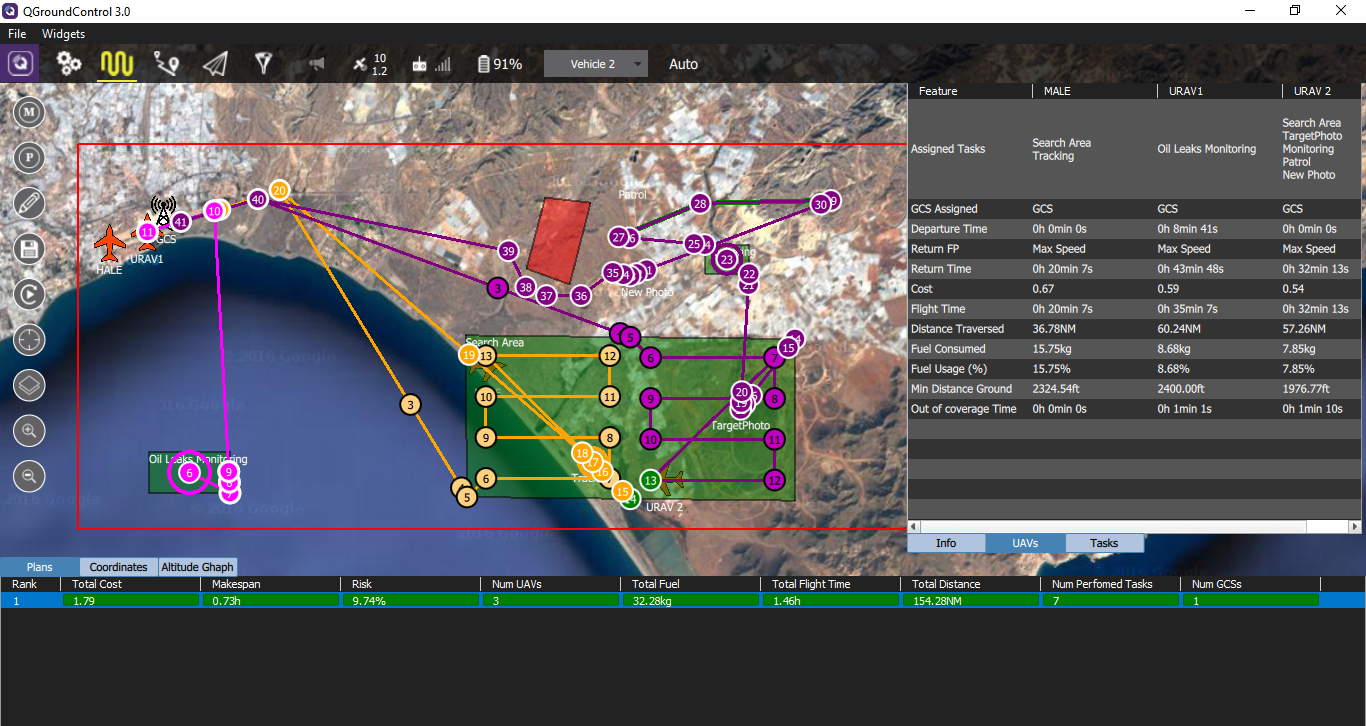}
\caption{Solution of the Mission Replanning.}\label{fig:mission_replanned}
\end{figure} 

\begin{figure}[H]
\centering
\includegraphics[width=\textwidth]{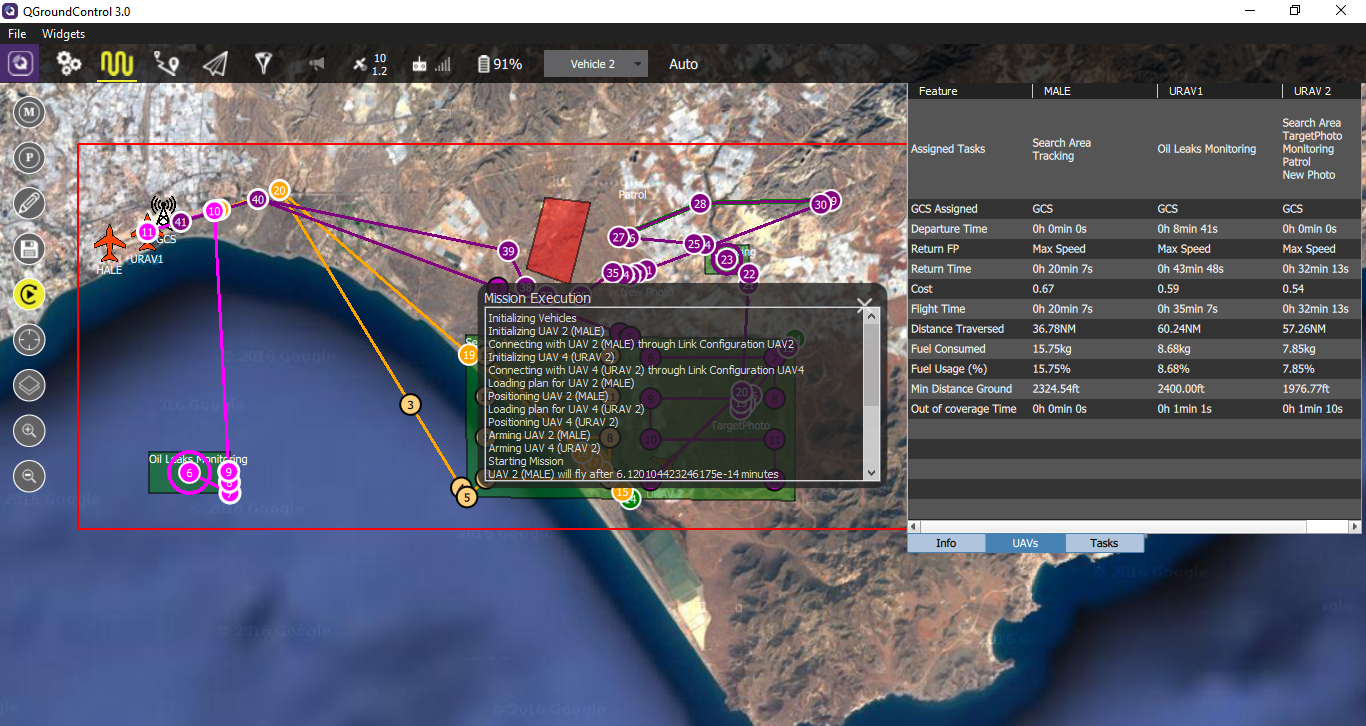}
\caption{Mission Update with the replanned solution.}\label{fig:mission_reexecution}
\end{figure} 

\begin{figure}[H]
\centering
\includegraphics[width=\textwidth]{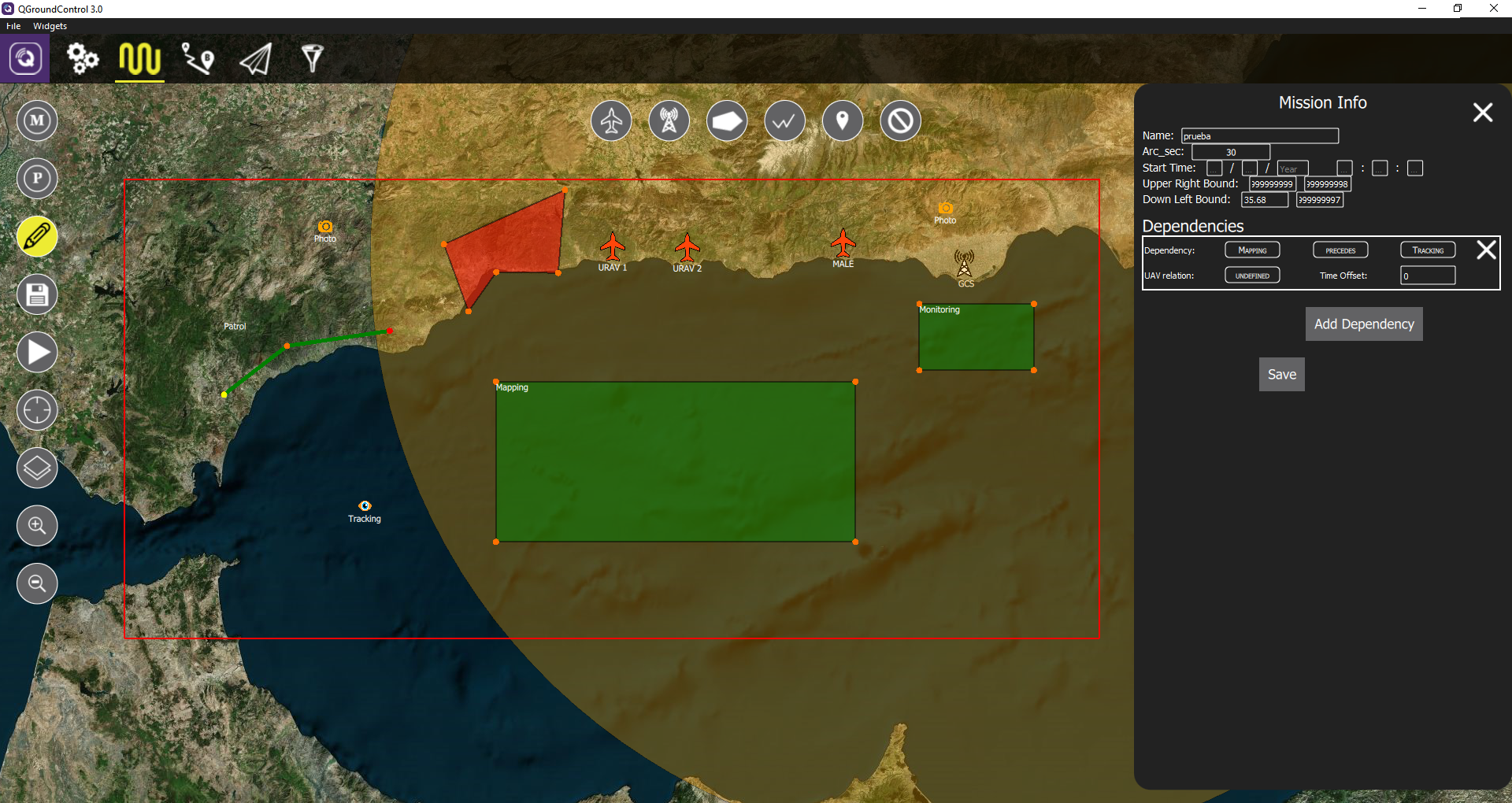}
\caption{Mission considered in the second use case.}\label{fig:case2}
\end{figure} 

\begin{figure}[H]
\centering
\includegraphics[width=\textwidth]{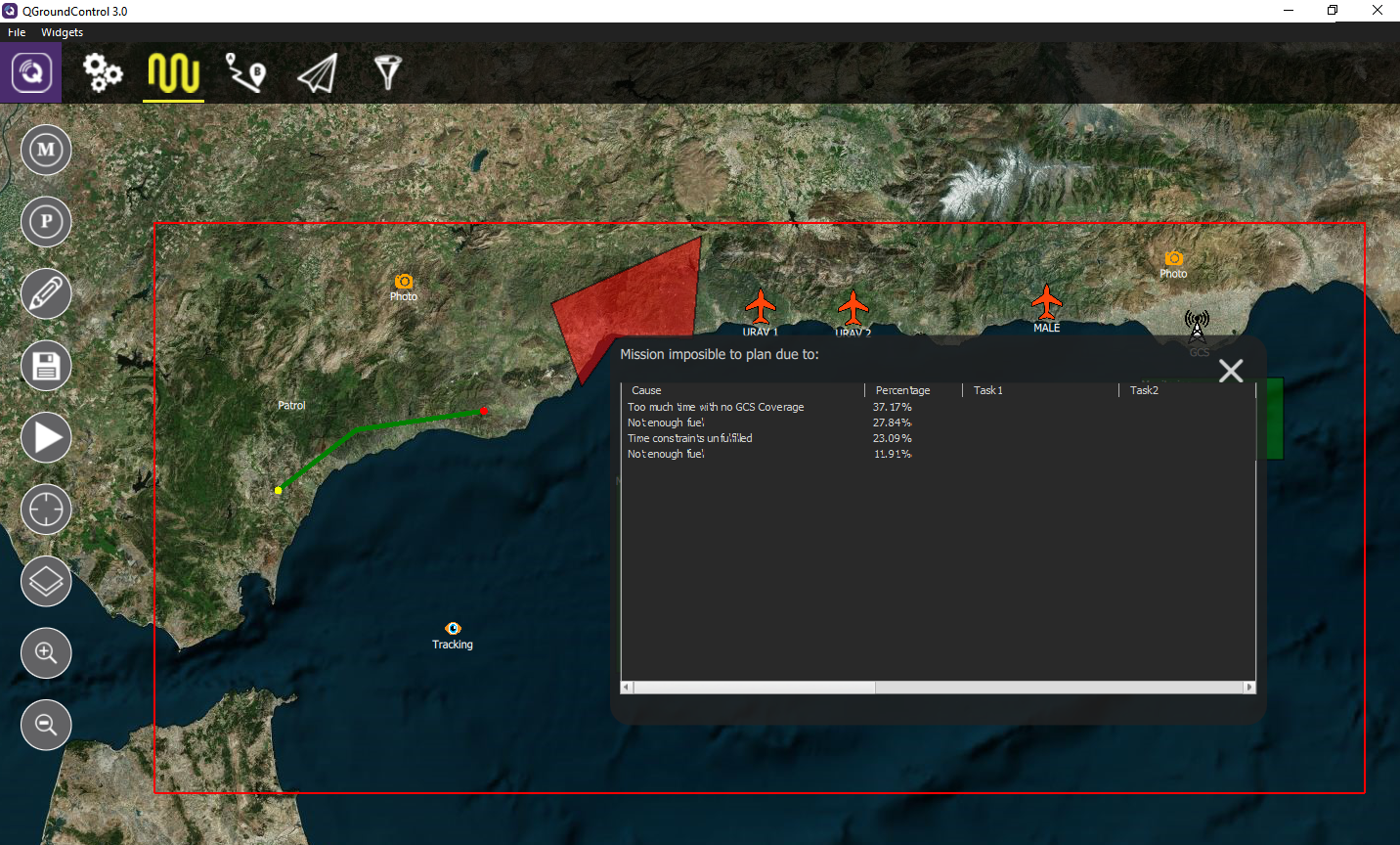}
\caption{No solutions obtained for the mission in the second use case.}\label{fig:case2_no_solutions}
\end{figure} 


\bibliographystyle{mdpi}


\bibliography{lite.bib}

\begin{thebibliography}{-------}
\providecommand{\natexlab}[1]{#1}

\bibitem[Kanistras \em{et~al.}(2015)Kanistras, Martins, Rutherford, and
  Valavanis]{Kanistras2015Survey}
Kanistras, K.; Martins, G.; Rutherford, M.J.; Valavanis, K.P., Survey of
  Unmanned Aerial Vehicles (UAVs) for Traffic Monitoring.
\newblock In {\em Handbook of Unmanned Aerial Vehicles}; Valavanis, K.P.;
  Vachtsevanos, G.J., Eds.; Springer Netherlands,  2015; pp. 2643--2666.

\bibitem[Zhang and Kovacs(2012)]{Zhang2012Application}
Zhang, C.; Kovacs, J.M.
\newblock The application of small unmanned aerial systems for precision
  agriculture: a review.
\newblock {\em Precision Agriculture} {\bf 2012}, {\em 13},~693--712.

\bibitem[Erdelj and Natalizio(2016)]{Erdelj2016UAV}
Erdelj, M.; Natalizio, E.
\newblock UAV-assisted disaster management: Applications and open issues.
\newblock  2016 International Conference on Computing, Networking and
  Communications (ICNC),  2016.

\bibitem[Evers \em{et~al.}(2014)Evers, Dollevoet, Barros, and
  Monsuur]{Evers2014Robust}
Evers, L.; Dollevoet, T.; Barros, A.I.; Monsuur, H.
\newblock Robust UAV mission planning.
\newblock {\em Annals of Operations Research} {\bf 2014}, {\em 222},~293--315.

\bibitem[Ram(2018)]{Ramirez2018Weighted}
{Weighted strategies to guide a multi-objective evolutionary algorithm for
  multi-UAV mission planning}.
\newblock {\em Swarm and Evolutionary Computation} {\bf 2018}, pp. 1--16.

\bibitem[Fukushima and Science(2012)]{Fukushima2012Onboard}
Fukushima, Y.; Science, A.
\newblock {Onboard Mission Replanning Using Operation Script and Orthogonal}.
\newblock  i-SAIRAS: International Symposium on Artificial Intelligence,
  Robotics and Automation in Space,  2012.

\bibitem[Ramirez-Atencia \em{et~al.}(2016)Ramirez-Atencia, Bello-Orgaz,
  R-Moreno, and Camacho]{Ramirez-Atencia2016MOGAMR}
Ramirez-Atencia, C.; Bello-Orgaz, G.; R-Moreno, M.D.; Camacho, D.
\newblock {MOGAMR: A Multi-Objective Genetic Algorithm for Real-Time Mission
  Replanning}.
\newblock  2016 IEEE Symposium Series on Computational Intelligence (SSCI).
  IEEE,  2016.

\bibitem[Ramirez-Atencia \em{et~al.}(2017)Ramirez-Atencia, Mostaghim, and
  Camacho]{Ramirez-Atencia2017Knee}
Ramirez-Atencia, C.; Mostaghim, S.; Camacho, D.
\newblock {A Knee Point based Evolutionary Multi-objective Optimization for
  Mission Planning Problems}.
\newblock  Genetic and Evolutionary Computation Conference (GECCO 2017). ACM,
  2017, pp. 1216--1223.

\bibitem[qgr(2018)]{qgroundcontrol}
{QGroundControl GCS}.
\newblock Available online: \url{http://www.qgroundcontrol.org/},  2018.

\bibitem[Sibley \em{et~al.}(2016)Sibley, Coyne, Avvari, Mishra, and
  Pattipati]{Sibley2016Supporting}
Sibley, C.; Coyne, J.; Avvari, G.V.; Mishra, M.; Pattipati, K.R.
\newblock Supporting Multi-objective Decision Making Within a Supervisory
  Control Environment.
\newblock  Foundations of Augmented Cognition: Neuroergonomics and Operational
  Neuroscience; Schmorrow, D.D.; Fidopiastis, C.M., Eds.,  2016, pp. 210--221.

\bibitem[Garcia and Barnes(2010)]{Garcia2010Multi}
Garcia, R.; Barnes, L.
\newblock Multi-UAV Simulator Utilizing X-Plane.
\newblock  2nd International Symposium on UAVs; Valavanis, K.P.; Beard, R.; Oh,
  P.; Ollero, A.; Piegl, L.A.; Shim, H., Eds.,  2010, pp. 393--406.

\bibitem[Rodriguez-Fernandez \em{et~al.}(2015)Rodriguez-Fernandez, Menendez,
  and Camacho]{Rodriguez-Fernandez2015Design}
Rodriguez-Fernandez, V.; Menendez, H.D.; Camacho, D.
\newblock Design and development of a lightweight multi-UAV simulator.
\newblock  2015 IEEE 2nd International Conference on Cybernetics (CYBCONF),
  2015, pp. 255--260.

\bibitem[ard(2018)]{ardupilot}
{ArduPilot}.
\newblock Available online: \url{http://ardupilot.org/},  2018.

\bibitem[px4(2018)]{px4}
{PX4 Pro autopilot}.
\newblock Available online: \url{http://px4.io/},  2018.

\bibitem[mav(2018)]{mavlink}
{MAVLink. Micro Air Vehicle Communication Protocol.}
\newblock Available online: \url{https://mavlink.io/},  2018.

\bibitem[mav(2017)]{mavproxy}
{MAVProxy}.
\newblock Available online:
  \url{https://ardupilot.github.io/MAVProxy/html/index.html},  2017.

\bibitem[mis(2018)]{missionplanner}
{Mission Planner}.
\newblock Available online: \url{http://ardupilot.org/planner/},  2018.

\bibitem[apm(2018)]{apmplanner2}
{APM Planner 2}.
\newblock Available online: \url{http://ardupilot.org/planner2/},  2018.

\bibitem[ugc(2018)]{ugcs}
{UgCS}.
\newblock Available online: \url{https://www.ugcs.com/},  2018.

\bibitem[pap(2018)]{paparazzi}
{Paparazzi UAV}.
\newblock Available online: \url{http://paparazziuav.org/},  2018.

\bibitem[Buro and Furtak(2003)]{Buro2003RTS}
Buro, M.; Furtak, T.
\newblock RTS games as test-bed for real-time AI research.
\newblock  Proceedings of the 7th Joint Conference on Information Science (JCIS
  2003),  2003, Vol. 2003, pp. 481--484.

\bibitem[Oliveira \em{et~al.}(2011)Oliveira, Cruz, Marques, and
  {a}o]{Oliveira2011Test}
Oliveira, T.; Cruz, G.; Marques, E.R.B.; {a}o, P.E.
\newblock A test bed for rapid flight testing of UAV control algorithms.
\newblock  Proc. Workshop on Research, Development and Education on Unmanned
  Aerial Systems (RED-UAS),  2011.

\bibitem[Triantaphyllou(200)]{Triantaphyllou2000}
Triantaphyllou, E.
\newblock {\em Multi-criteria Decision Making Methods: A Comparative Study};
  Springer US,  200; p. 290.

\bibitem[Deb \em{et~al.}(2002)Deb, Pratap, Agarwal, and Meyarivan]{Deb2002Fast}
Deb, K.; Pratap, A.; Agarwal, S.; Meyarivan, T.
\newblock A fast and elitist multiobjective genetic algorithm: {NSGA-II}.
\newblock {\em Evolutionary Computation} {\bf 2002}, {\em 6},~182--197.

\bibitem[Branke \em{et~al.}(2004)Branke, Deb, Dierolf, and
  Osswald]{Branke2004Finding}
Branke, J.; Deb, K.; Dierolf, H.; Osswald, M.
\newblock {Finding knees in multi-objective optimization}.
\newblock  Parallel Problem Solving from Nature - PPSN VIII. PPSN 2004. Lecture
  Notes in Computer Science; Yao, X., Ed. Springer, Berlin, Heidelberg,  2004,
  Vol. 3242, pp. 722--731.

\bibitem[Schulte \em{et~al.}(2010)Schulte, Tack, and
  Lagerkvist]{Schulte2010Modeling}
Schulte, C.; Tack, G.; Lagerkvist, M.Z.
\newblock {Modeling and Programming with Gecode}.
\newblock Available online: \url{http://www.gecode.org/},  2010.

\bibitem[Opricovic and Tzeng(2004)]{Opricovic2004Compromise}
Opricovic, S.; Tzeng, G.H.
\newblock Compromise solution by MCDM methods: A comparative analysis of VIKOR
  and TOPSIS.
\newblock {\em European Journal of Operational Research} {\bf 2004}, {\em
  156},~445 -- 455.

\end{thebibliography}

\sampleavailability{Samples of the compounds ...... are available from the authors.}

\end{document}